%% file: main.tex
\documentclass[10pt,journal,compsoc]{IEEEtran}
\usepackage[hidelinks]{hyperref}

\ifCLASSOPTIONcompsoc
  \usepackage[nocompress]{cite}
\else
  \usepackage{cite}
\fi
\ifCLASSINFOpdf
\else
\fi
\usepackage[pdftex]{graphicx}

\usepackage{amsmath}
\usepackage{amsfonts}
\usepackage{lmodern}

\usepackage{cuted}
\usepackage{capt-of}
\usepackage{stfloats}

\usepackage[utf8]{inputenc}

\usepackage{subcaption}
\begin{document}
\bstctlcite{IEEEexample:BSTcontrol}
\title{Visualizing Movement Control Optimization Landscapes}


\author{Perttu~Hämäläinen,
        Juuso~Toikka,
		Amin~Babadi,
        and~C.~Karen~Liu% <-this % stops a space
\IEEEcompsocitemizethanks{\IEEEcompsocthanksitem Hämäläinen,  
Toikka, and Babadi are with the Department of Computer Science at Aalto University.\protect\\
E-mail: perttu.hamalainen@aalto.fi, juuso.toikka@aalto.fi, amin.babadi@aalto.fi
\IEEEcompsocthanksitem C. Karen Liu is with the Department of Computer Science
at Stanford University.\protect\\
E-mail: karenliu@cs.stanford.edu}% <-this % stops an unwanted space
%\thanks{Manuscript received November 20, 2019}
}


\markboth{Accepted to IEEE Transactions on Visualization and Computer Graphics}%
{Hämäläinen \MakeLowercase{\textit{et al.}}: Visualizing Movement Control Optimization Landscapes}
\IEEEtitleabstractindextext{%

\begin{abstract}
  \input{abstract}
\end{abstract}

}

% \twocolumn[{%
% \renewcommand\twocolumn[1][]{#1}%
\maketitle

\IEEEdisplaynontitleabstractindextext

% \fi
% \IEEEpeerreviewmaketitle

\input{body}

\newpage

\begin{IEEEbiography}[{\includegraphics[width=1in,height=1.25in,clip,keepaspectratio]{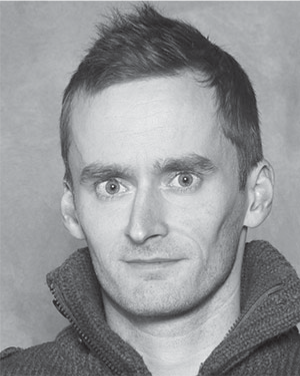}}]{Perttu Hämäläinen}
  received an M.Sc.(Tech) degree from Helsinki University of Technology in 2001, an M.A. degree from the University of Art and Design Helsinki in 2002, and a doctoral degree in computer science from Helsinki University of Technology in 2007. Presently, Hämäläinen is an associate professor at Aalto University, publishing on human-computer interaction, computer animation, machine learning, and game research. Hämäläinen is passionate about human movement in its many forms, ranging from analysis and simulation to first-hand practice of movement arts such as parkour or contemporary dance.
\end{IEEEbiography}

\begin{IEEEbiography}[{\includegraphics[width=1in,height=1.25in,clip,keepaspectratio]{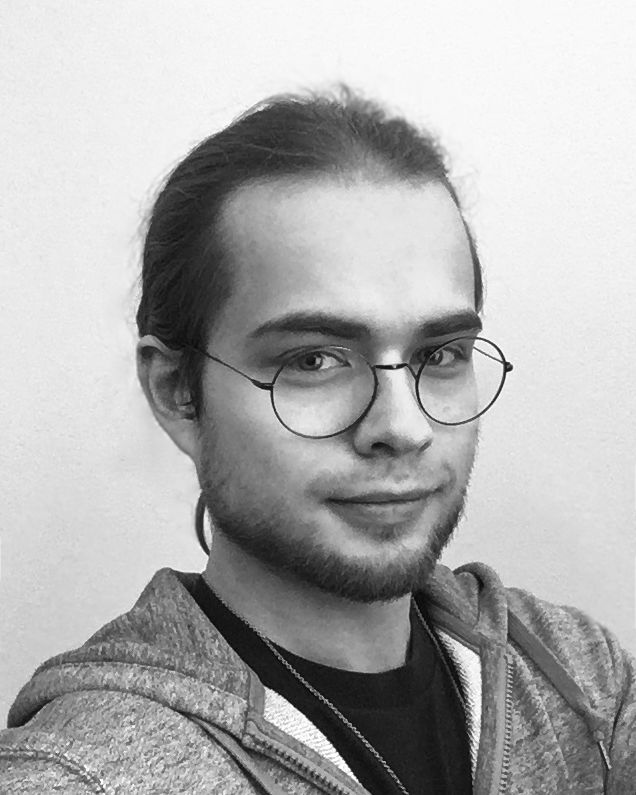}}]{Juuso Toikka}
  received his M.Sc.(Tech) degree in Computer Science from Aalto University in 2019. While this manuscript was in preparation, Toikka worked as a research assistant at the Department of Computer Science at Aalto University, Finland, but he has since moved on to Ubisoft RedLynx, pursuing a game industry career. His professional interests include animation tools, procedural animation, movement control optimization, reinforcement learning, and emergence in games.
\end{IEEEbiography}

\begin{IEEEbiography}[{\includegraphics[width=1in,height=1.25in,clip,keepaspectratio]{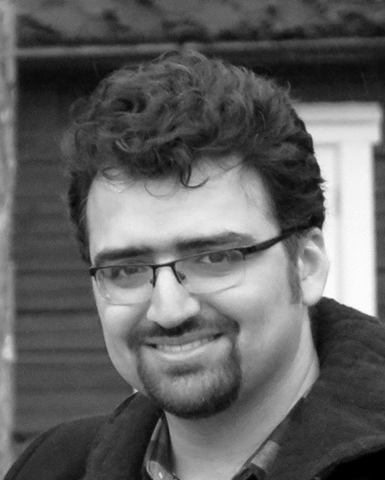}}]{Amin Babadi}
 is a doctoral candidate at the Department of Computer Science, Aalto University, Finland. His research focuses on developing efficient, creative movement artificial intelligence for physically simulated characters in multi-agent settings. Babadi has previously worked on three commercial games, developing AI, animation, gameplay, and physics simulation systems.
\end{IEEEbiography}
% \begin{IEEEbiographynophoto}{Juuso Toikka}

%\newpage

\begin{IEEEbiography}[{\includegraphics[width=1in,height=1.25in,clip,keepaspectratio]{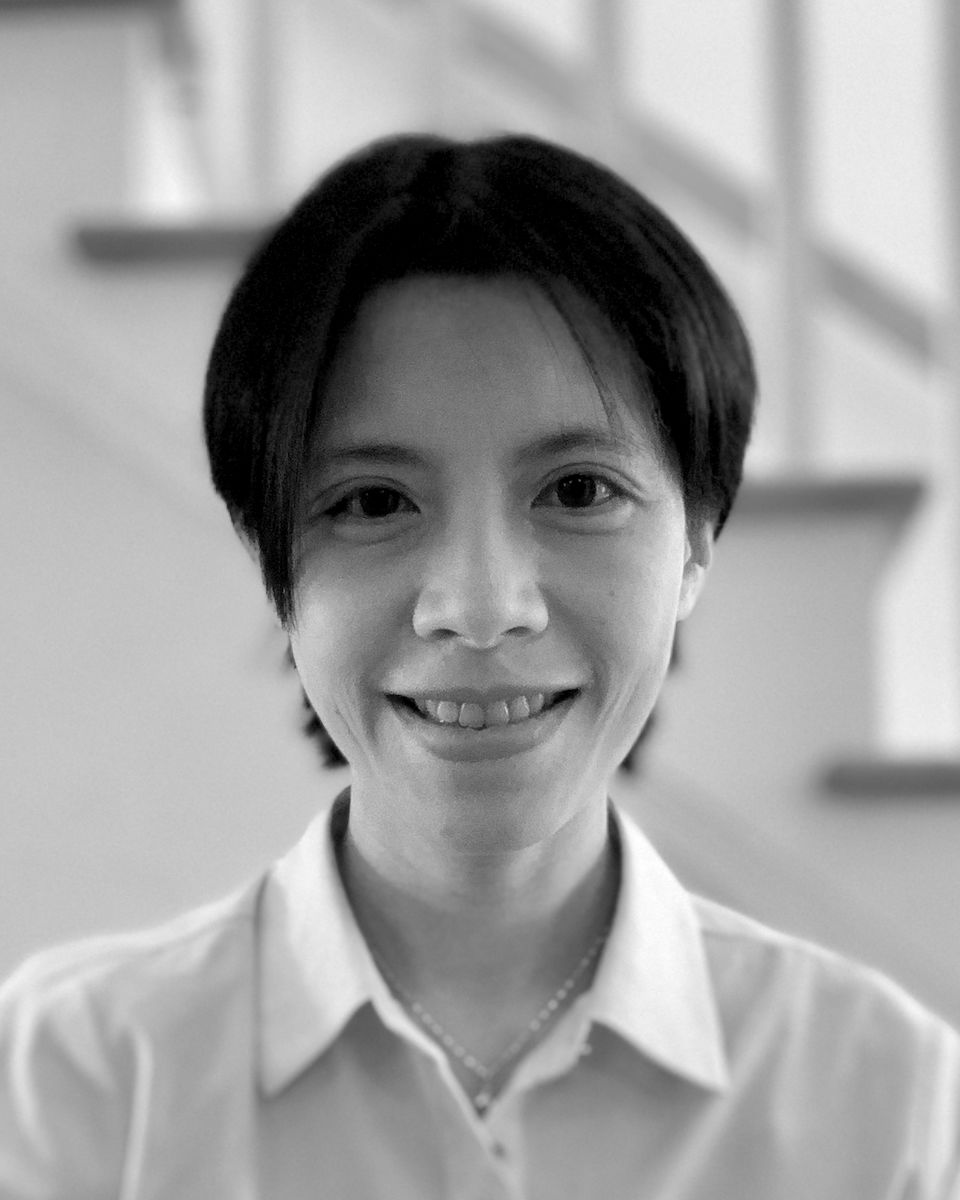}}]{C. Karen Liu}
  is an associate professor in the Department of Computer Science at Stanford University. She received her Ph.D. degree in Computer Science from the University of Washington. Liu's research interests are in computer graphics and robotics, including physics-based animation, character animation, optimal control, reinforcement learning, and computational biomechanics. She developed computational approaches to modeling realistic and natural human movements, learning complex control policies for humanoids and assistive robots, and advancing fundamental numerical simulation and optimal control algorithms. The algorithms and software developed in her lab have fostered interdisciplinary collaboration with researchers in robotics, computer graphics, mechanical engineering, biomechanics, neuroscience, and biology. Liu received a National Science Foundation CAREER Award, an Alfred P. Sloan Fellowship, and was named Young Innovators Under 35 by Technology Review. In 2012, Liu received the ACM SIGGRAPH Significant New Researcher Award for her contribution in the field of computer graphics.
\end{IEEEbiography}


%\vfill


%\appendix[Paper Supplement: Visualizing Movement Control Optimization Landscapes]
\twocolumn[\section*{Paper Supplement: Visualizing Movement Control Optimization Landscapes}]
\begin{figure*}[b]
\centering
{\includegraphics[width=0.9\linewidth]{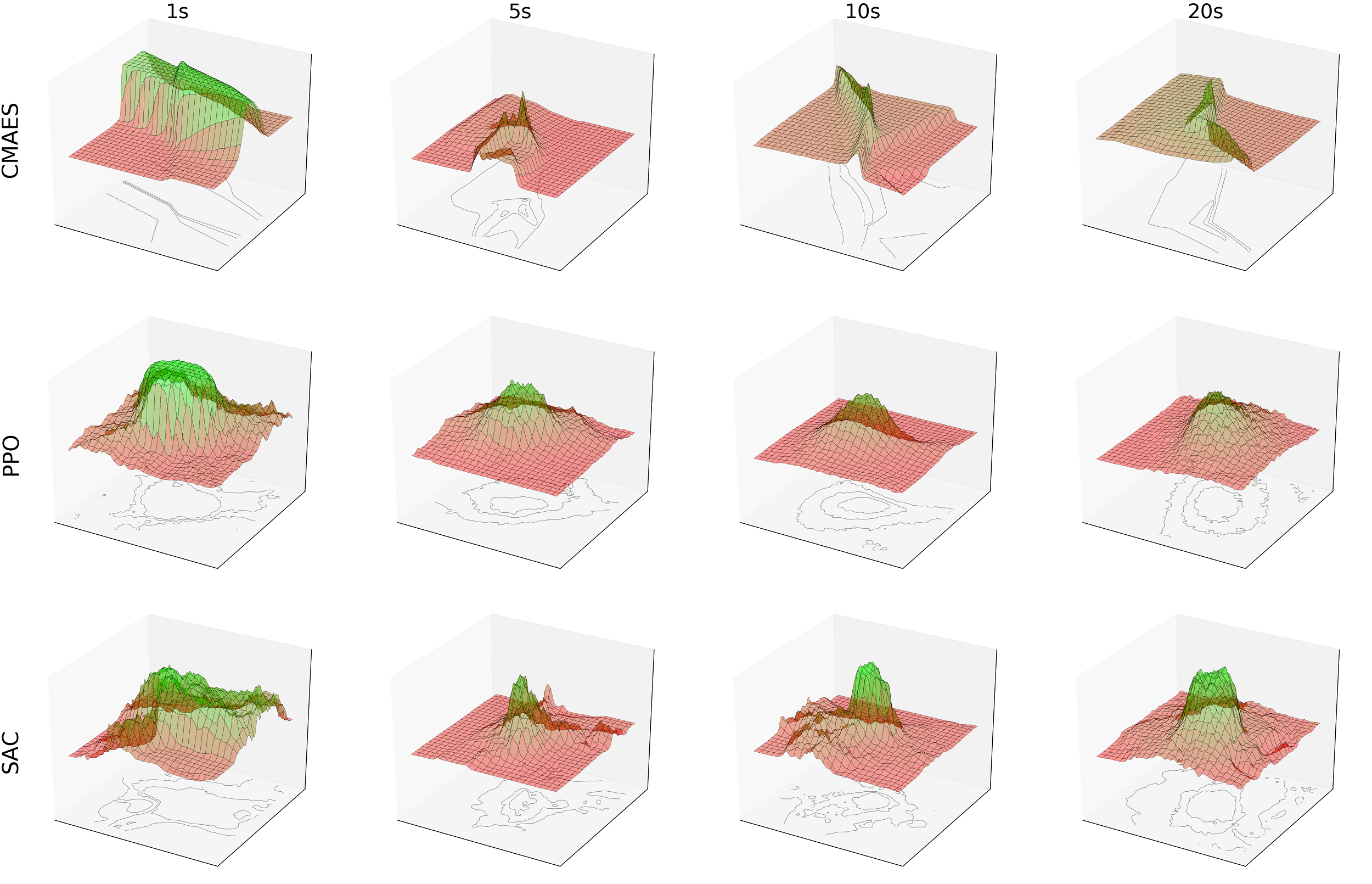}}
\caption{Trajectory optimization (CMA-ES) and policy optimization (PPO, SAC) landscapes for the Hopper-v2 MuJoCo environment, with trajectory/episode lengths ranging from 1 to 20 seconds. Trajectory optimization becomes highly ill-conditioned for long trajectories.}
\label{fig:landscape_hopper}
\end{figure*}

This supplementary document presents additional results to augment the paper's Section 9.4. Recall that the central result of Section 9.4 is that policy optimization scales better to long trajectories/episodes, although a policy neural network typically has orders of magnitude more parameters to optimize than a single trajectory (even a long one). This was tested with multiple locomotion tasks and optimizers: A Unity Machine Learning Agents 3D humanoid (optimized using LM-MA-ES and PPO) and four different MuJoCo agents: A 2D monopedal hopper (Hopper-v2), 2D bipedal walker (Walker2d-v2), 2D half quadruped (HalfCheetah-v2), and a 3D humanoid (Humanoid-v2), optimized using CMA-ES, PPO, and SAC. The optimization landscape visualizations of section 9.4---from the Unity humanoid locomotion case---agree on the result, displaying much less multimodality and ill-conditioning in the policy optimization case. 

Similar landscape plots of the MuJoCo agents are included below. All landscape visualizations are centered around the found optima (a vector of control torques for each time step in trajectory optimization, or a vector of neural network parameters in policy optimization). The visualizations were computed using grids of $100\times100$ points, computing the mean return of 10 trajectories/episodes for each grid point. To improve visual clarity, all landscapes were also filtered using Gaussian blur with $\sigma=1.0$.

These MuJoCo landscapes support the results of Section 9.4. This is clearest in the hopper landscapes shown in Fig.~\ref{fig:landscape_hopper}. CMA-ES trajectory optimization landscapes become increasingly ill-conditioned with long trajectories, with narrow ridges where an optimizer typically zigzags back and forth over the ridge, making very slow progress along the ridge. In contrast, the policy optimization landscapes show almost spherical optima. Note that the policy networks for PPO and SAC have different parameter counts (as per the default parameters of the Stable Baselines implementations that we used). Thus, the landscapes cannot display exactly same optima. 

\begin{figure*}[!ht]
\centering
{\includegraphics[width=0.9\linewidth]{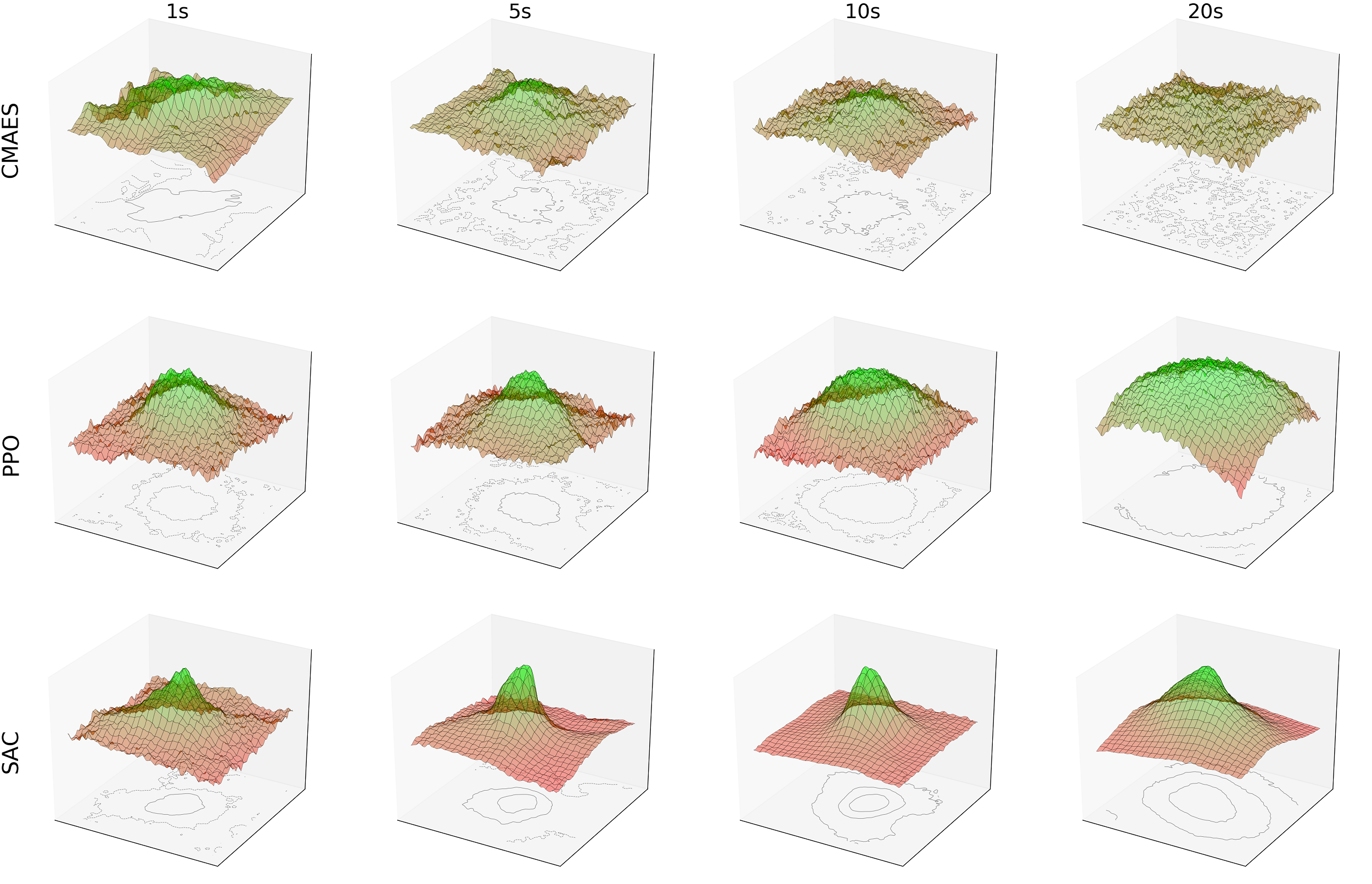} }
\caption{Trajectory optimization (CMA-ES) and policy optimization (PPO, SAC) landscapes for the HalfCheetah-v2 MuJoCo environment, with trajectory/episode lengths ranging from 1 to 20 seconds.}
\label{fig:landscape_halfcheetah}
\end{figure*}

\begin{figure*}[!ht]
\centering
{\includegraphics[width=0.9\linewidth]{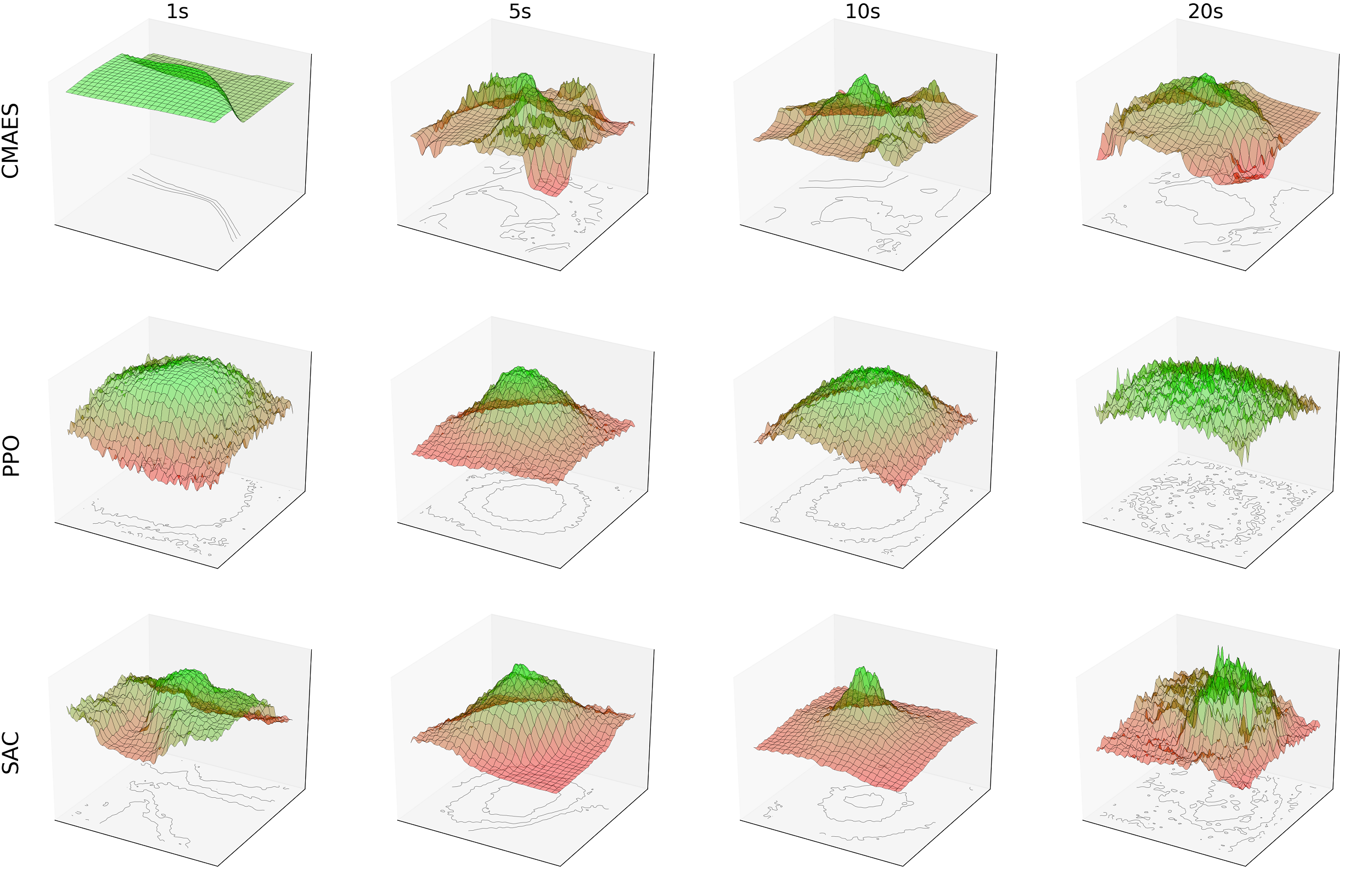} }
\caption{Trajectory optimization (CMA-ES) and policy optimization (PPO, SAC) landscapes for the Humanoid-v2 MuJoCo environment, with trajectory/episode lengths ranging from 1 to 20 seconds.}
\label{fig:landscape_humanoid}
\end{figure*}

The plots for the other MuJoCo environments (Fig. \ref{fig:landscape_halfcheetah}-\ref{fig:landscape_humanoid}) exhibit similar qualities, although less clearly. The trajectory optimization landscapes also become increasingly multimodal and/or noisy with longer trajectories. It should be noted that each landscape's vertical axis is normalized to show maximal detail, i.e., the heights of the optima in different landscapes cannot be directly compared.

\begin{figure*}[!ht]
\centering
{\includegraphics[width=0.9\linewidth]{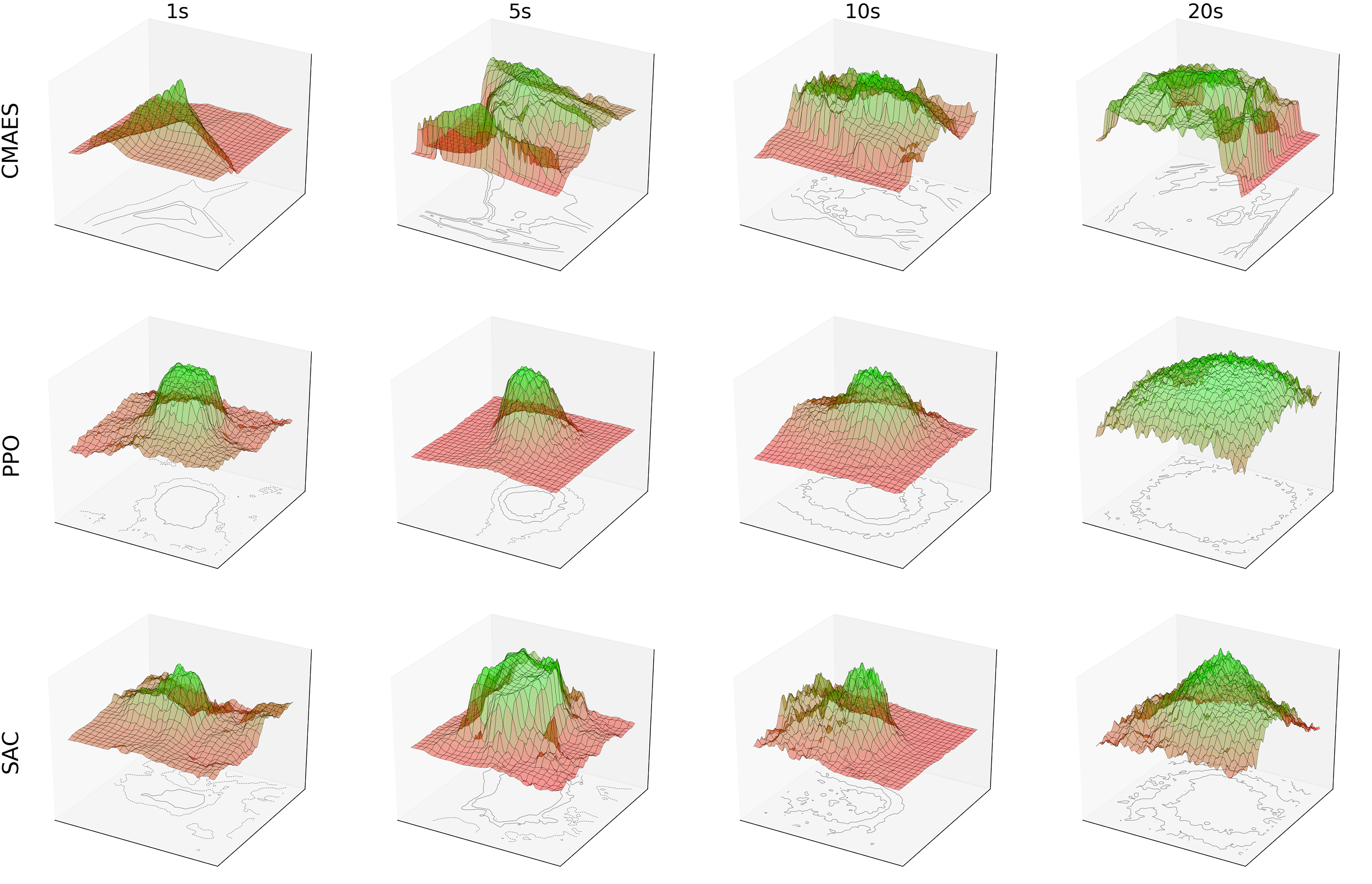} }
\caption{Trajectory optimization (CMA-ES) and policy optimization (PPO, SAC) landscapes for the Walker2d-v2 MuJoCo environment, with trajectory/episode lengths ranging from 1 to 20 seconds.}
\end{figure*}

\end{document}

%% file: abstract.tex
%!TEX root = main.tex

A large body of animation research focuses on optimization of movement control, either as action sequences or policy parameters. However, as closed-form expressions of the objective functions are often not available, our understanding of the optimization problems is limited. Building on recent work on analyzing neural network training, we contribute novel visualizations of high-dimensional control optimization landscapes; this yields insights into why control optimization is hard and why common practices like early termination and spline-based action parameterizations make optimization easier. For example, our experiments show how trajectory optimization can become increasingly ill-conditioned with longer trajectories, but parameterizing control as partial target states---e.g., target angles converted to torques using a PD-controller---can act as an efficient preconditioner. Both our visualizations and quantitative empirical data also indicate that neural network policy optimization scales better than trajectory optimization for long planning horizons. Our work advances the understanding of movement optimization and our visualizations should also provide value in educational use.  

%Our visualizations also illustrate how early termination can remove local optima in both trajectory and policy optimization, but combining negative rewards with termination may result in the emergence of false optima.

%% file: body.tex
%!TEX root = main.tex
\IEEEraisesectionheading{\section{Introduction}\label{sec:introduction}}
\IEEEPARstart{M}{uch} of computer animation research formulates animation as an optimization problem. It has been shown that complex movements can emerge from minimizing a cost function that measures the divergence from movement goals, such as moving to a specific pose or location while minimizing effort. In principle, this holds the promise of elevating an animator to the role of a choreographer, directing virtual actors and stuntmen through the definition of movement goals. However, solving the optimization problems can be hard in practice. It can require hours or even days of computing time, which is highly undesirable for interactive applications and the rapid iteration of movement goals; defining the goals can be a non-trivial design problem in itself, requiring multiple attempts to produce a desired aesthetic result.

\begin{figure}[h!]
\centering
\includegraphics[width=2.9in]{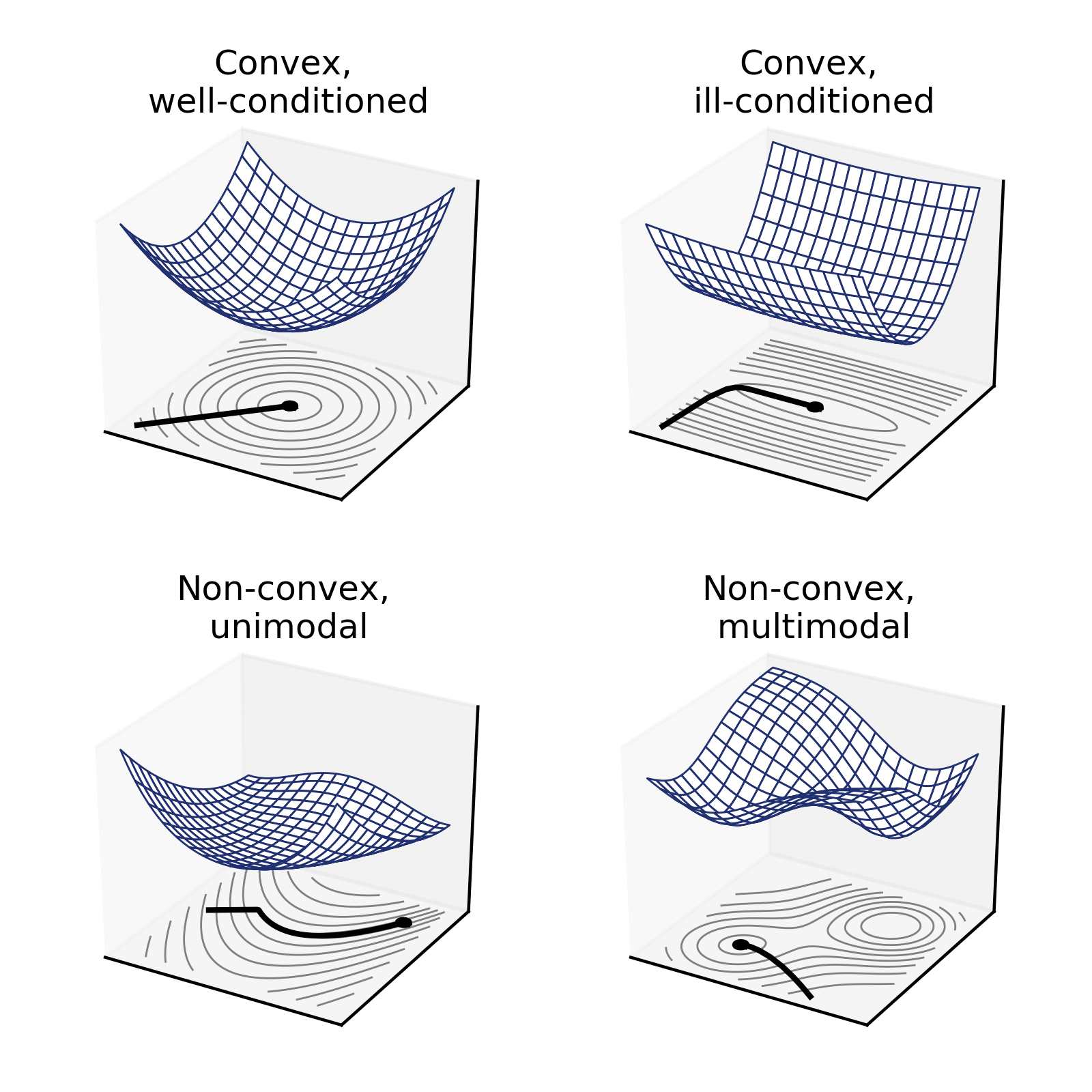}
\caption{Common types of optimization landscapes. The surfaces denote the values of 2-dimensional (bivariate) objective functions, with the isocontours displayed below the surface. The black curves show the progress of gradient descent optimization from an initial point.} \label{fig:introfigure}
\end{figure}

%Optimization problems can be divided into the four classes of increasing difficulty illustrated in Fig. \ref{fig:introfigure}. In general, one would like the objective function to be \textit{more convex, well-conditioned, and unimodal}. Convexity refers to the shape of the isocontours of a 2D objective function, or level sets in a general $d$-dimensional case. In a well-conditioned case, the isocontours are spherical instead of elongated, and simple gradient descend recovers a direct path to the optimum, as the gradients coincide with the isocontour normals.
Optimization problems can be divided into the four classes of increasing difficulty illustrated in Fig. \ref{fig:introfigure}:
\begin{itemize}
\item \textit{Convex and well-conditioned} Convexity refers to the shape of the isocontours of the objective function, or level sets in a general $d$-dimensional case. In the ideal well-conditioned case, the isocontours are spherical, and simple gradient descent recovers a direct path to the optimum.
\item \textit{Convex and ill-conditioned} In ill-conditioned optimization, the isocontours are elongated instead of spherical.  The gradient---coinciding with isocontour normals---no longer points towards the optimum, and numerical optimization may require more iterations. %\footnote{Second-order methods utilizing the Hessian of the objective function can avoid this, but are too computationally expensive for many real-world applications.}
\item \textit{Non-convex and unimodal} Non-convexity tends to make optimization even harder, but numerical optimization usually still converges if there are no local optima to distract it.
\item \textit{Non-convex and multimodal} In this problem class, the landscape has local optima which can attract optimization. Gradient-free, sampling-based approaches like CMA-ES---common in animation research---may still find the global optimum \cite{hansen2004evaluating}, but this can be computationally expensive. Unfortunately, movement optimization can easily fall into this class, e.g. due to the discontinuities caused by colliding objects, and multiple options for going around obstacles \cite{Hamalainen2014,Hamalainen2015}.
\end{itemize}

%\begin{figure*}[t]
%\centering
%\includegraphics[width=0.85\textwidth]{images/trajectory_teaser2d_wireframe.png}
%%\includegraphics[width=7.0in]{images/sweep_T_torques.png}
%\caption{Visualizing the effect of different problem modifications on the trajectory optimization landscape of an inverted pendulum, using a trajectory length of 100 timesteps, i.e., a 100-dimensional search space. See Section \ref{sec:problems} for details. From left to right, 1) "baseline" version where the pendulum is actuated with torques and using squared costs as the objective, 2) converting the cost $c_t$ of timestep $t$ to a reward through exponentiation as $r_t=e^{-c_t}$, 3) the same with early termination, which effectively removes local optima, and 4) same as the baseline, but parameterizing the control as a sequence of target angles, which makes the optimization considerably more separable and well-conditioned.}\label{fig:teaser}% 5) the same but with rewards instead of costs.}
%\end{figure*}

Landscape visualizations like those in Fig. \ref{fig:introfigure} provide useful intuitions of optimization problems, and can help in reformulating a problem into a more tractable form. In general, one would like the objective function to be \textit{more convex, well-conditioned, and unimodal}. We know that problem modifications such as the choice of action space can greatly affect movement optimization efficiency \cite{peng2017learning}, but visualizing the effects on the optimization landscape is challenging because of high problem dimensionality. High dimensionality and/or non-differentiable physics simulators can also prevent analyzing problem conditioning through the eigenvalues of the Hessian.  % and black box physics simulators that do not provide analytic Hessians.  %and visualizing the effects of the modifications on the optimization landscape could provide valuable insights into why they work so well. Unfortunately, such visualization cannot be directly applied to high-dimensional problems. 

The primary inspiration of this paper comes from recent work on visualizing neural network loss function landscapes by Li et al. \cite{li2018visualizing}. Strikingly, the paper shows that \textit{visualization of random 2D slices of a high-dimensional objective function can convey useful intuitions and predict the difficulty of optimization}, even with highly complex networks with millions of parameters.  More specifically, the approach generates 3D landscape plots of the objective function $f(\mathbf{x}): \mathbb{R}^d \rightarrow \mathbb{R}$ by evaluating it along a plane (a 2D subspace in $\mathbb{R}^d$) defined by two random orthogonal basis vectors intersecting the optimum. Li et al. \cite{li2018visualizing} use the approach to illustrate how deeper networks have more local optima, but adding skip-connections greatly helps in making the landscape more convex and unimodal.

%\textbf{Contribution:} We contribute by showing that the random slice visualization approach of Li et al. \cite{li2018visualizing} can be applied in the domain of movement optimization.Fig. \ref{fig:teaser} provides examples of this, using the same basis vectors in all the plots for more consistent comparisons. Fig. \ref{fig:randomslices} shows the same objective function with different random basis vectors. Furthermore, we use the visualizations to investigate the following research questions:

\begin{figure*}[!t]
\centering
\includegraphics[width=7.0in]{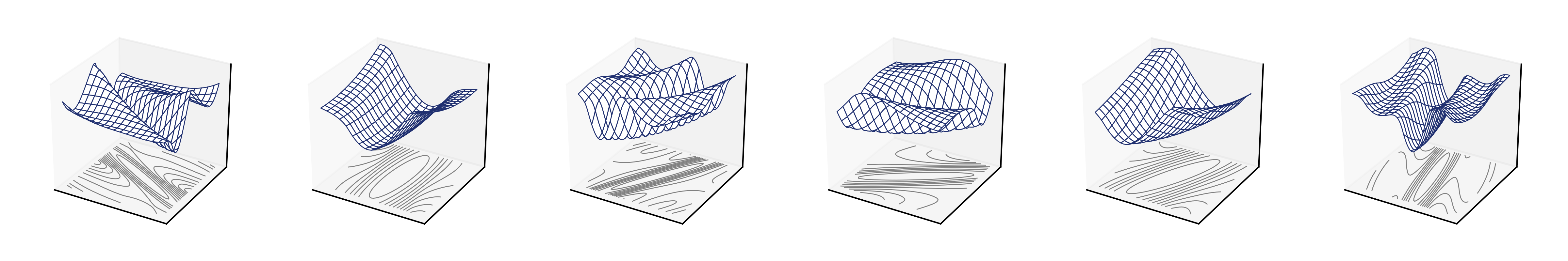}
\caption{Visualizing the inverted pendulum trajectory optimization objective of Equation \ref{eq:trajcost} with different random basis vectors and $T=100$. Although the plots are not exactly similar, they all exhibit the same overall structure, i.e., multimodality and an elongated, ill-conditioned optimum.}\label{fig:randomslices}
\end{figure*}

\textbf{Contribution:} We contribute by showing that the random slice visualization approach of Li et al. \cite{li2018visualizing} can be applied in the domain of movement optimization. Fig. \ref{fig:randomslices} shows examples of this, visualizing the same objective function with different random basis vectors. Furthermore, we use the visualizations to investigate the following research questions:
\begin{itemize}
\item What is the effect of the number of timesteps -- i.e., the planning horizon -- on the optimization landscape? (Section \ref{sec:timesteps})
\item What is the effect of the choice of action space on the optimization landscape? (Section \ref{sec:actionspace})
\item What is the effect of converting instantaneous costs into rewards through exponentiation? (Section \ref{sec:rewards})
\item What is the effect of early termination of movement trajectories or episodes, e.g., when deviating from a target state? (Section \ref{sec:termination})
\item Do the visualizations predict actual optimization performance, and generalize from simple to complex problems? (Sections \ref{sec:optcompare} and \ref{sec:biped}).
\end{itemize}

Additionally, Section \ref{sec:theory} provides a more theoretical investigation of the reliability and limitations of random 2D slice visualizations. We conclude that such visualization is a useful tool for diagnosing problems; the somewhat low sensitivity is  compensated by high specificity. Our visualizations also explain why movement optimization best practices such as early termination and parameterizing actions as target angles work so well, which should make our work useful in teaching computer animation and movement optimization.

%On the other hand, information is inevitably lost in the 2D visualizations. Appendix \ref{sec:theory} analyzes how the visualizations can fail to show ill-conditioning or multimodality, although our empirical results do not demonstrate such problems. To be on the safe side, we complement our discussion with analysis of the Hessian and direct plots of simple 1D policy optimization objective functions.

\section{Related work}
\noindent\textbf{Spacetime optimization} Much of the earlier work on animation as optimization focused on extensions of the seminal work on spacetime optimization by Witkin and Kass \cite{witkin_spacetime_1988,cohen_interactive_1992,fang_efficient_2003,safonova_synthesizing_2004,wampler2009optimal}, where the optimized variables included the root position and rotation as well as joint rotations for each animation frame. However, the synthesized motions were limited by the need for prior knowledge of contact information, such as when and which body parts should make contact with the ground. This limitation was overcome by \cite{mordatch_discovery_2012}, who introduced auxiliary optimized variables that specify the contact information. However, the number of colliding body parts was still limited.

\par\smallskip
\noindent\textbf{Animation as simulation control} In recent years, the focus of research has shifted towards animation as a simulation control problem. Typically, one optimizes simulation control parameters such as time-varying actuation torques of character joints, and an off-the-shelf physics simulator is used to realize the movement. While spacetime optimization can be performed with gradient-based optimization methods like Sequential Quadratic Programming \cite{witkin_spacetime_1988} or L-BFGS \cite{mordatch_discovery_2012}, simulation control is commonly approached with sampling-based, gradient-free optimization due to non-differentiable dynamics and/or multimodality \cite{Liu2010,Hamalainen2014,Hamalainen2015}. This is also what our work focuses on; it remains as future work to extend our visualizations to analyzing spacetime optimization.

\par\smallskip
\noindent\textbf{Trajectory and policy optimization} Two main classes of approaches include trajectory and policy optimization. In trajectory optimization, one optimizes the time-varying control parameters directly, which can be done both offline \cite{ngo_spacetime_1993,al_borno_trajectory_2013,naderi2017discovering} or online, while the character moves and acts \cite{tassa2012synthesis,Hamalainen2014,Hamalainen2015}. In policy optimization, one optimizes the parameters of a policy function such as a neural network that maps character state to (approximately) optimal control, typically independent of the current simulation time. This can be done both using neuroevolution \cite{Geijtenbeek2013,such2017deep} or Reinforcement Learning (RL), which has recently proven powerful even with complex humanoid movements \cite{schulman2017proximal,peng2018deepmimic,lee2019scalable,bergamin2019drecon,park2019learning}. Unfortunately policy optimization/learning can be computationally expensive with large neural networks, and may require careful curriculum design \cite{yu2018learning}. On the other hand, it can produce controllers that require orders of magnitude less computing resources after training, compared to using trajectory optimization to solve each required movement in an interactive application such as a video game. Trajectory and policy optimization approaches can also be combined \cite{levine2013guided,mordatch2014combining,rajamaki2018continuous}, which allows one to adjust the trade-off between training time and runtime expenses. In this paper, we provide analyses of both trajectory and policy optimization landscapes.

\par\smallskip

\noindent\textbf{Visualizing optimization} Many optimization visualizations are problem-specific, utilizing the semantics of optimized parameters \cite{jones1994visualization}. Visualization is also used for letting a user interact and inform optimization \cite{meignan2015review}; this, however, falls outside the scope of this paper. Considering non-interactive, generic methods applicable to continuous-valued optimization, landscape visualizations like the ones in Fig. \ref{fig:introfigure} are a standard textbook method. Although it is technically trivial to extend this to higher-dimensional problems by visualizing the objective function on a random plane (a 2D subspace), Li et al. \cite{li2018visualizing} only recently demonstrated that such random slices can provide meaningful insights and, likewise, have some predictive power on the difficulty of very high-dimensional optimization. Inspired by \cite{li2018visualizing}, we test the random slice visualization approach in a new domain, and also provide additional analyses of the method's reliability and limitations. Other common methods for visualizing high-dimensional optimization include graphing the objective function along a straight line from the initial point to the found optimum \cite{goodfellow2014qualitatively,keskar2016large,dinh2017sharp,smith2017exploring}, or visualizing in a plane determined from the path taken during optimization \cite{goodfellow2014qualitatively}. There are also examples of visualizing movement optimization through a conversion to an interactive game or puzzle; in this case, players perform the optimization aided by predictive visualizations of how different actions affect the simulation state \cite{hamalainen2017predictive}. 

In computer animation and movement control research, objective functions are visualized occasionally, using various approaches to reduce the objectives to 2D. For example, Hämäläinen et al. \cite{Hamalainen2014} visualize contact discontinuities and multimodality in a 2D toy problem, and Sok et al. \cite{sok2007simulating} visualize a high-dimensional multimodal objective with respect to two manually selected parameters. However, we know of no previous paper that focuses on visualizing movement optimization, or applies the 2D random slice approach of Li et al. \cite{li2018visualizing} to movement optimization.

\section{Test Problem: Inverted Pendulum Balancing}\label{sec:problems}
\begin{figure}[h]
\centering
\includegraphics[width=1.0in]{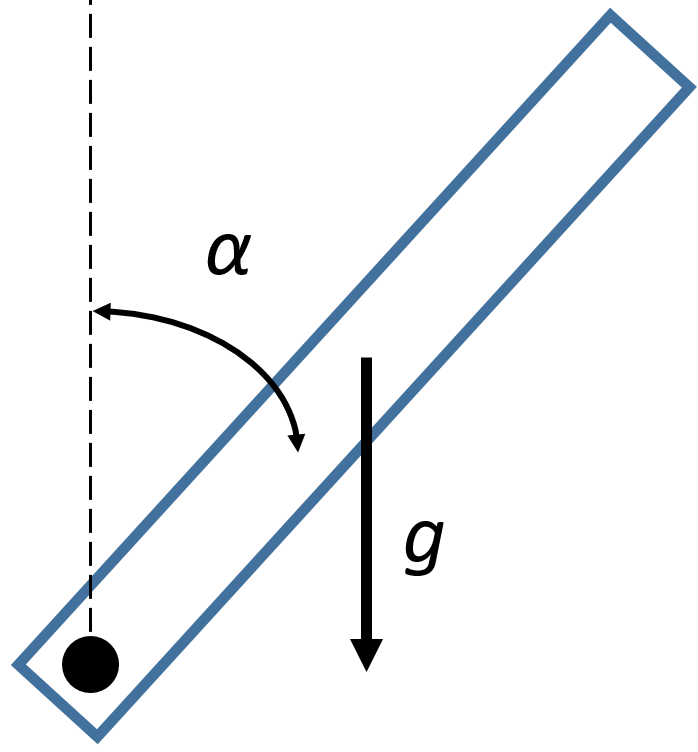}
\caption{The inverted pendulum model. The force exerted by gravity is denoted by $g$, and $\alpha$ denotes the angular deviation from an upright position.}\label{fig:pendulum}
\end{figure}

This section describes the inverted pendulum balancing problem that is used throughout Sections \ref{sec:timesteps}-\ref{sec:optcompare}, before testing the visualization approach on the more complex simulated humanoid of Section \ref{sec:biped}. The pendulum is depicted in Fig. \ref{fig:pendulum}. Although it is simple, it offers the following benefits for analyzing movement optimization:
  
\begin{itemize}
\item In the trajectory optimization case, we know the true optimum. As the pendulum dynamics are differentiable, we can also compute the Hessian and its eigenvalues for further analysis. We implement this using Autograd \cite{maclaurin2015autograd}.
\item In the policy optimization case, we can use a simple P-controller as the parameteric policy, which admits visualizing the full optimization landscape instead of only low-dimensional slices.
\end{itemize}

%To test whether the insights gained from the inverted pendulum are applicable to more complex problems,   However, in this case the true optimum is not known, and we are limited to visualizing the landscape in around an empirically found near-optimal trajectory.

The dynamics governing the angle $\alpha_t$, angular velocity $\omega_t$, and control torque $\tau_t$ of the pendulum at timestep $t$ are implemented as:

\begin{eqnarray}
\omega_t &=& \omega_{t-1} + \delta (\tau_t + 0.5\ l\ g \sin(\alpha_{t-1})),\label{eq:avelUpdate}\\
\alpha_t &=& \alpha_{t-1} + \delta \omega_t,
\end{eqnarray}
where $\delta$, $l$, and $g$ are the simulation timestep, pendulum length, and force induced by gravity, respectively. We use $\delta=0.1$, $l=0.2$, and $g=0.981$.

\subsection{Trajectory Optimization}\label{sec:trajopt}
Most of our trajectory optimization visualizations are generated from the problem of balancing a simple inverted pendulum, starting from an upward position such that the optimal torque sequence $\tau_1, ..., \tau_T$ is all zeros. The subscripts denote timestep indices and $T$ is the planning horizon, i.e., the length of the simulated trajectory. The optimization objective is to minimize the trajectory cost $\mathcal{C}$ computed as the sum of instantaneous costs:

\begin{equation}
\mathcal{C}=\sum_{t=1}^T (\alpha_t^2 + w \tau_t^2). \label{eq:trajcost} \\ 
\end{equation}
The cost is minimized when the pendulum stays upright at $\alpha=0$ with zero torques. The relative importance of state cost $\alpha_t^2$ and action cost $\tau_t^2$ is adjusted by the multiplier $w$. We use $w=1$ unless specified otherwise. The cost landscape is visualized in case of $T=100$ in Fig. \ref{fig:randomslices}.

Some of our experiments convert the cost minimization problem into a reward maximization problem, computing trajectory reward $\mathcal{R}$ using exponentiated costs as

%\mathcal{R}=\sum_{t=1}^T e^{-(\alpha_t^2 + w \tau_t^2)}.\label{eq:trajreward}
\begin{equation}
\mathcal{R}=\sum_{t=1}^T ( e^{-\alpha_t^2} + w e^{-\tau_t^2} ).\label{eq:trajreward}
\end{equation}
%Although the cost minimization and reward maximization problems are not exactly the same, Section \ref{sec:rewards} will show that they have similar structure. 
The reward formulation has been recently used with stellar results in the policy optimization of complex humanoid movements \cite{peng2018deepmimic}. Exponentiation is also used in framing optimal control as estimation \cite{todorov2008general,todorov2009efficient}.

\subsection{Policy Optimization}
In the case of policy optimization, we use the same pendulum simulation, with a minor adjustment. Instead of directly optimizing control torques, we use a policy $\tau_t = \pi_\theta (\mathbf{s}_t)$, parametrized by $\theta$, where $\mathbf{s}$ denotes pendulum state. The optimization objective is to either minimize the expected trajectory cost, or maximize the expected reward, assuming that each trajectory or ``episode'' is started from a random initial state. As closed-form expressions for the expectations are not available, we replace them by averages. These are computed from 10 episodes, each started from a different initial pendulum angle. 

In all the pendulum policy optimization visualizations, we use a simple P-controller as the policy:  

\begin{equation}
\pi_\theta (\mathbf{s}_t) = \theta \alpha_t.
\end{equation}

%Unlike in trajectory optimization, we do not need to initialize with zero angle and angular velocity, as due to the simple policy, we can plot the full landscape Compared to the zero-angle initialization of the trajectory optimization case, we use this more standard %In the figures of this paper, $\mathcal{C}$ or $\mathcal{R}$ are used to denote either the average episode cost or reward.

\begin{figure}[th]
\centering
\includegraphics[width=\linewidth]{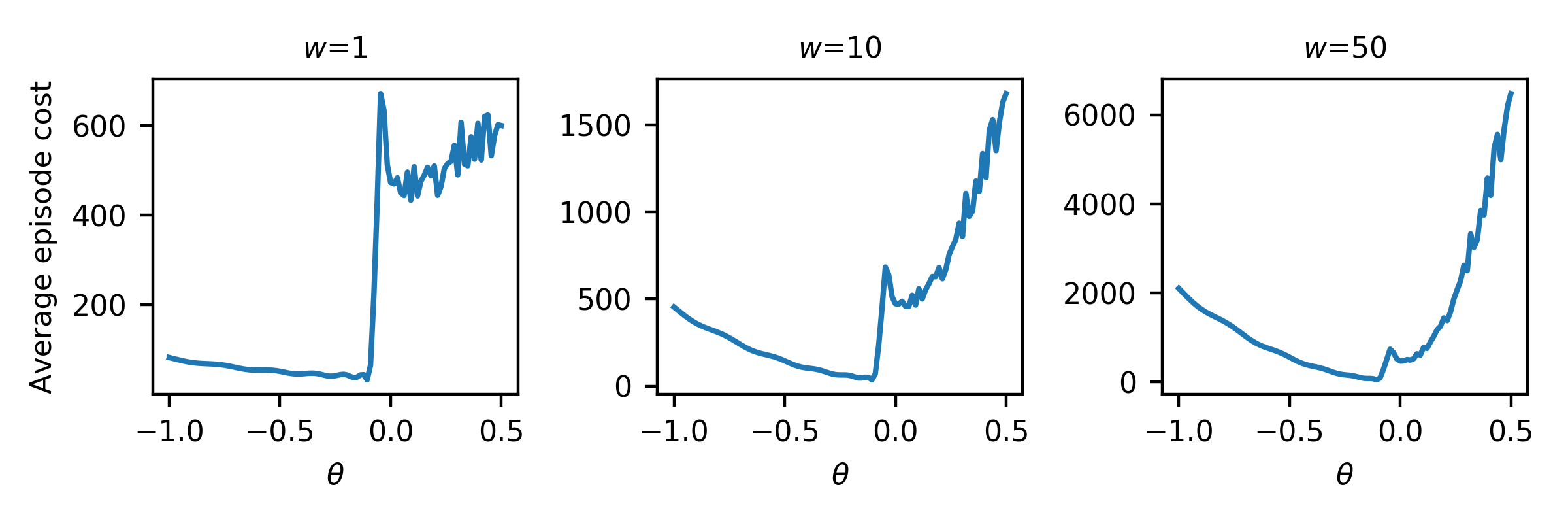}
\caption{Average episode cost (Equation \ref{eq:trajcost}) with $T=200$, as a function of the P-controller policy parameter $\theta$, and action cost weight $w$. A large $w$ makes the local and true optima more equally good compared to the surrounding regions. This makes it more likely that a global Monte Carlo optimization method like CMA-ES will get attracted to the false optimum. The success of local gradient-based optimization depends on which cost basin the optimization is initialized in.} \label{fig:policy}
\end{figure}

The benefit of this formulation is that we only have a single policy parameter to optimize; thus, we can visualize the full objective function, shown in Fig. \ref{fig:policy}. Despite the simplicity, this provides a multimodal optimization problem with properties similar to more complex problems. As shown in Fig. \ref{fig:policy}, there is global optimum at approximately $\theta=-0.1$, and also a false optimum near $\theta=0$. At the false optimum, the action cost is minimized simply by letting the pendulum hang downwards. In a real-world case like controlling a simulated humanoid, the false optimum corresponds to resting on the ground with zero effort; readers familiar with humanoid control probably know that such a behavior is easy to elicit by having too large an effort cost. %Note that there's also some noise due to positive $\theta$ accelerating the pendulum to a very high angular velocity that makes the simulated states somewhat random with our large timestep. 

%=======================================================================

\section{Effect of trajectory length} \label{sec:timesteps}
\begin{figure}[th]
\centering
\includegraphics[width=\linewidth]{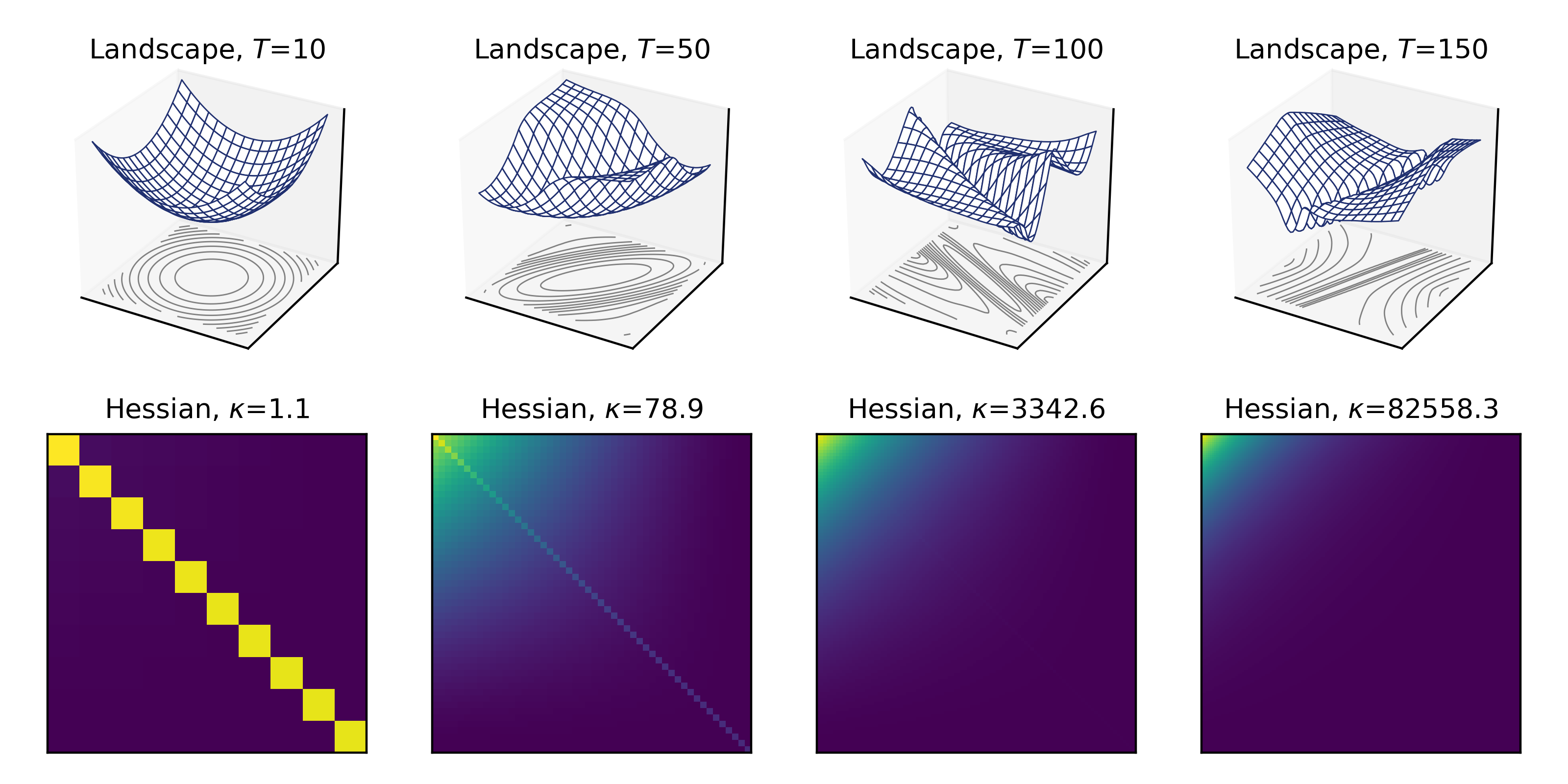}
\caption{Effect of trajectory length $T$ on inverted pendulum trajectory optimization. The optimization problem becomes increasingly ill-conditioned and non-separable with longer action sequences. The bottom row shows the Hessian matrices at the optimal points with increasing $T$. $\kappa$ denotes the condition number of the Hessian.  }\label{fig:sweepT}
\end{figure}

\begin{figure}[th]
\centering
\includegraphics[width=\linewidth]{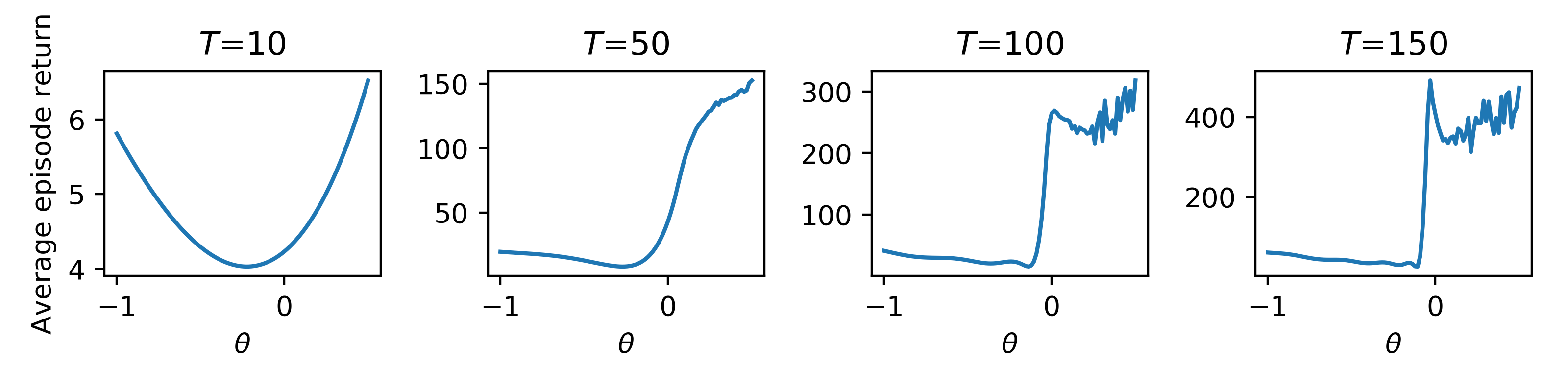}
\caption{Effect of trajectory length $T$ on inverted pendulum trajectory policy optimization. Local optima become pronounced with large $T$, as the agent has more time to diverge and accumulate cost far from the desired states.}\label{fig:sweepT_policy}
\end{figure}

Figures \ref{fig:sweepT} and \ref{fig:sweepT_policy} visualize the inverted pendulum trajectory and policy optimization landscapes, with different trajectory and episode lengths $T$. The figures yield two main insights:

\begin{itemize} 
\item{Trajectory optimization can become increasingly  non-separable and ill-conditioned with large $T$.}

\item{In both trajectory and policy optimization, the landscapes become more multimodal with large $T$.} 
\end{itemize}

%With long action sequences, even small perturbations of initial actions require corrective later actions, to avoid a large deviation of the state trajectory.

\subsection{Trajectory Optimization} 
\label{sec:horizon_trajectory}
The trajectory optimization landscapes of Fig. \ref{fig:sweepT} are augmented with visualizations of the Hessian matrices of the cost function at the optimum. This allows further analysis of some important properties. A diagonal Hessian means that the optimization problem is separable, and the variables can be optimized independent of each others. Strong off-diagonal elements imply that if one changes a variable, then one must also change some other variable to remain at the bottom of the valley in the landscape. 

On the other hand, the eigenvalues of the Hessian measure curvature along the eigenvectors; the condition number $\kappa$, which denotes the ratio of the largest to smallest eigenvalues of the Hessian, is generally considered as a predictor of optimization difficulty. In the ideal case, the Hessian is a (scaled) identity matrix, i.e., a diagonal matrix where all the eigenvalues are the same; this indicates both separability and no ill-conditioning with $\kappa=1$.

Intuitively, the ill-conditioning with a large $T$ can be explained by the fact that perturbing an action of the optimal trajectory will lead to state divergence that accumulates over time. Thus, the total state cost is more sensitive to earlier actions, which leads to large differences in the Hessian eigenvalues. Non-separability stems from a need to adjust later actions to correct the state divergence. On the other hand, with a small $T$, the state has less time to diverge, and the $\tau^2$ action cost dominates; this gives rise to the constant diagonal structure of the Hessian, as actions of all timesteps contribute equally and independently to the cost. 

More formally, we may rewrite the cost function (Equation \ref{eq:trajcost}) of the trajectory optimization as:
\begin{equation}
    \mathcal{C}=\sum_{t=1}^T \alpha_t(\tau_1, \cdots, \tau_t)^2 + \sum_{t=1}^T w \tau_t^2,  \\ 
\end{equation}
emphasizing that the state $\alpha_t$ depends on the past actions $\tau_1$ to $\tau_t$. The Hessian of the cost function can then be expressed as:
\begin{equation}
    \frac{\partial^2 \mathcal{C}}{\partial \boldsymbol{\tau}^2}=2 \sum_{t=1}^T \big(\nabla_{\boldsymbol{\tau}} \alpha_t (\nabla_{\boldsymbol{\tau}} \alpha_t)^T + \alpha_t \frac{\partial^2 \alpha_t }{\partial \boldsymbol{\tau}^2}\big) + 2w \mathbf{I}.  \\ 
    \label{eq:cost_hessian}
\end{equation}

The last term of Equation \ref{eq:cost_hessian} is a diagonal matrix that does not depend on $T$, while the first term results in off-diagonal elements accumulated over time. As $T$ grows larger, the first term begins dominating the second one, hence the decrease in separability (i.e., the fading diagonal shown in Fig. \ref{fig:sweepT}). 

The increase of the condition number follows from the causality of the dynamics---future actions do not affect past states. This means that when $t=1$, the first term of Equation \ref{eq:cost_hessian} is a $T \times T$ matrix that has only one nonzero element at the upper-left corner, while, when $t=T$, the nonzero elements occupy the entire matrix. Summing up all the matrices from $t=1$ to $t=T$, the first term of the Hessian will have larger values toward the upper-left corner, as evidenced by the bright areas in Fig. \ref{fig:sweepT}. The longer $T$ is, the more disparate the upper-left and bottom-right corners become, leading to larger condition numbers.

Although it is well known that the difficulty of solving a trajectory optimization increases proportionally with the length of the planning horizon, our visualization shows that the difficulty is not solely caused by the increase in the size of the problem, but also by the increasingly ill-conditioned and non-separable objective function. Section \ref{sec:actionspace} explains how the choice of action parameters can act as an efficient preconditioner.

\subsection{Policy Optimization}
In policy optimization, the optimized parameters have no similar dependency to the horizon $T$, as the effect of each policy parameter on the objective is averaged over all states. The separability of policy parameters depends on the function representation of the policy. For example, if the policy is represented as a multi-layer nonlinear neural network, the Hessian of the objective function is bound to be inseparable, as the neuron weights across layers are multiplied with each other when passing data through the network.

However, a longer horizon in policy optimization can induce a multimodal landscape in the objective function (Fig. \ref{fig:sweepT_policy}). An explanation is that a longer time budget enables strategies that are not possible when the horizon is shorter. The ``mollification'' of the landscape at shorter horizons explains why policy optimization may benefit from a curriculum in which the planning horizon is gradually increased. The OpenAI Five Dota 2 bots provide a recent impressive example of this, gradually increasing the $\gamma$ parameter of Proximal Policy Optimization \cite{schulman2017proximal} during training\footnote{https://blog.openai.com/openai-five/}.

%=======================================================================

\section{Effect of the Choice of Action Space} \label{sec:actionspace}
In the previous section, we saw that minimizing effort as the sum of squared actions gives rise to a strong constant diagonal in the Hessian, which in principle leads to better conditioned optimization. Parameterizing actions as torques, and having a squared torque cost term, is also common in continuous control benchmark problems such as OpenAI Gym MuJoCo \cite{brockman2016openai}. However, \cite{peng2017learning} showed that parameterizing actions as target joint angles can make policy optimization more effective. The target poses that the policy outputs are converted to joint torques, typically using a P- or PD-controller. Such pose-based control is also common in earlier work on trajectory optimization \cite{al2013trajectory,Hamalainen2014,Hamalainen2015,Rajamaki:2017:ASB:3099564.3099579}, often claimed to give better results than optimizing raw torques; despite this, comprehensive comparisons of control parameterizations are rare. 

\begin{figure}[t]
\centering
\includegraphics[width=\linewidth]{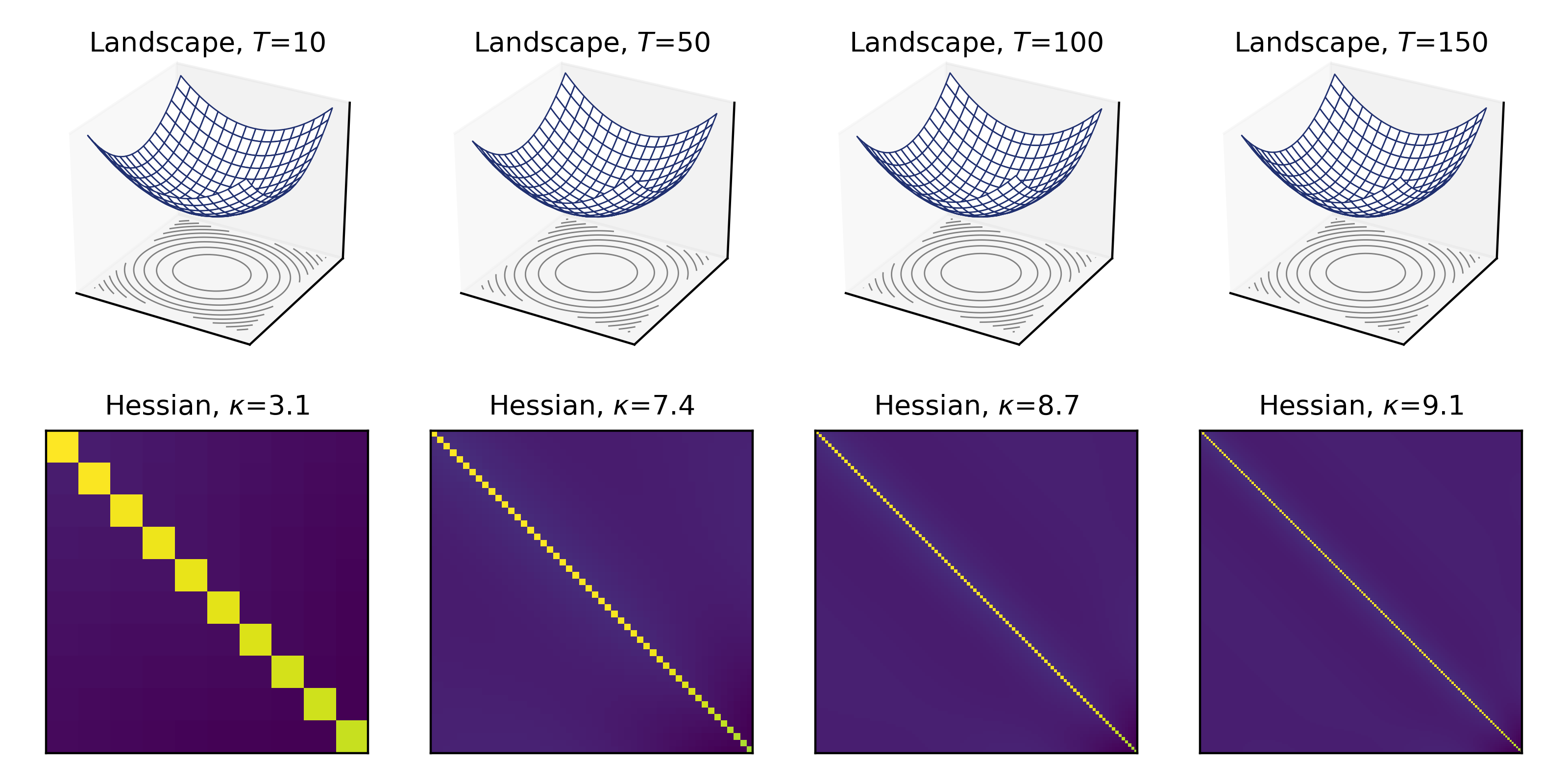}
\caption{Effect of episode length $T$ on inverted pendulum trajectory optimization, when parameterizing actions as target angles. Remarkably, as opposed to the torque parameterization of Fig. \ref{fig:sweepT}, the landscape becomes much less ill-conditioned with large $T$.}\label{fig:sweepT_angles}
\end{figure}

Fig. \ref{fig:sweepT_angles} shows that using actions as target angles $\bar{\alpha}$ does indeed scale better to long horizons. The strong constant diagonal of the Hessian does not fade out as much, and the condition number $\kappa$ grows much slower with larger $T$. We compute the torques using a PD-controller: $\tau = k_p (\bar{\alpha}-\alpha) + k_d \omega$, using $k_p=1$ and $k_d=-1$. The optimal $\bar{\alpha}$ sequence is all zeros, as the pendulum starts from an upright position with zero velocity.

An explanation for the effectiveness of the target angle parameterization is that \textit{actions represent (partial) target states}. This makes the optimization of state cost or reward terms more separable and well-conditioned, and the Hessian closer to a scaled identity matrix. The optimal action at each timestep is more independent of preceding actions; a reasonable strategy is to always drive the pendulum towards the desired $\alpha=0$. This explains why the Hessian is closer to diagonal. Furthermore, with the state-based control, perturbing early actions leads to less cumulative state divergence; hence, all actions contribute more equally to the cost. This explains why the spread of the diagonal values is low. 

It should be noted that when using target joint angles with a character with an unactuated root, the parameterization cannot fully remove the dependencies between the actions of different timesteps. Deviations in initial actions can still lead to divergence, e.g. falling out of balance, which needs to be corrected by later actions.

We can also see the effect of action choice from the dynamics of the pendulum. Using torques as actions directly, the dynamic equation for the angle can be expressed as:
\begin{equation}
\alpha_t = \alpha_{t-1} + \delta \omega_{t-1} + \delta^2 (\tau_t + 0.5\ l\ g \sin(\alpha_{t-1})),
\label{eq:torque_action}
\end{equation}
where the dependency to past states and actions is primarily due to the first two terms. When using target angles as actions, we replace $\tau_t$ with $\bar{\alpha}_{t-1} - \alpha_{t-1} - \omega_{t-1}$ and arrive at a slightly different dynamic equation:
\begin{equation}
    \alpha_t = (1 - \delta^2) \alpha_{t-1} + (\delta - \delta^2)\omega_{t-1} + \delta^2 (\bar{\alpha}_{t-1} + 0.5\ l\ g \sin(\alpha_{t-1})).
\label{eq:angle_action}
\end{equation}

Now, the previous angle $\alpha_{t-1}$ and velocity $\omega_{t-1}$ have smaller multipliers $(1-\delta^2)$ and $(\delta - \delta^2)$. This discount is applied recursively over time, such that the angle and velocity at $n$ time steps ago are discounted by $(1-\delta^2)^n$ and $(\delta-\delta^h)^n$. Using the same reasoning of causality as described in Section \ref{sec:horizon_trajectory}, the exponential reduction in dependency on previous states makes the Hessian of the cost function better conditioned (Equation \ref{eq:cost_hessian}).

\subsection{Pose splines}\label{sec:splines}
In many papers, long action sequences are parameterized as "pose splines", i.e. parameteric curves that define the target pose of the controlled character over time, which are then implemented using P- or PD-controllers \cite{al2013trajectory,Hamalainen2014,naderi2017discovering}. The discussion above provides a strong motivation for this, as such parameterization achieves two things at once: 
\begin{itemize}
\item Better separability and conditioning of the problem due to the choice of action space. 
\item Improved avoidance of ill-conditioning with long action sequences; in effect, the control points of a spline can be thought of as a shorter sequence of higher-level actions, each of which defines the instantaneous actions for multiple timesteps. 
\end{itemize}

Naturally, using splines has aesthetic motivations as well, as they result in smoother motion with less frame-by-frame noise. 

Experimental results comparing spline-based optimization with direct optimization of action sequences are provided in Section \ref{sec:biped}.

%=======================================================================

\section{Effect of using rewards instead of costs} \label{sec:rewards}
\begin{figure}[t]
\centering
\includegraphics[width=\linewidth]{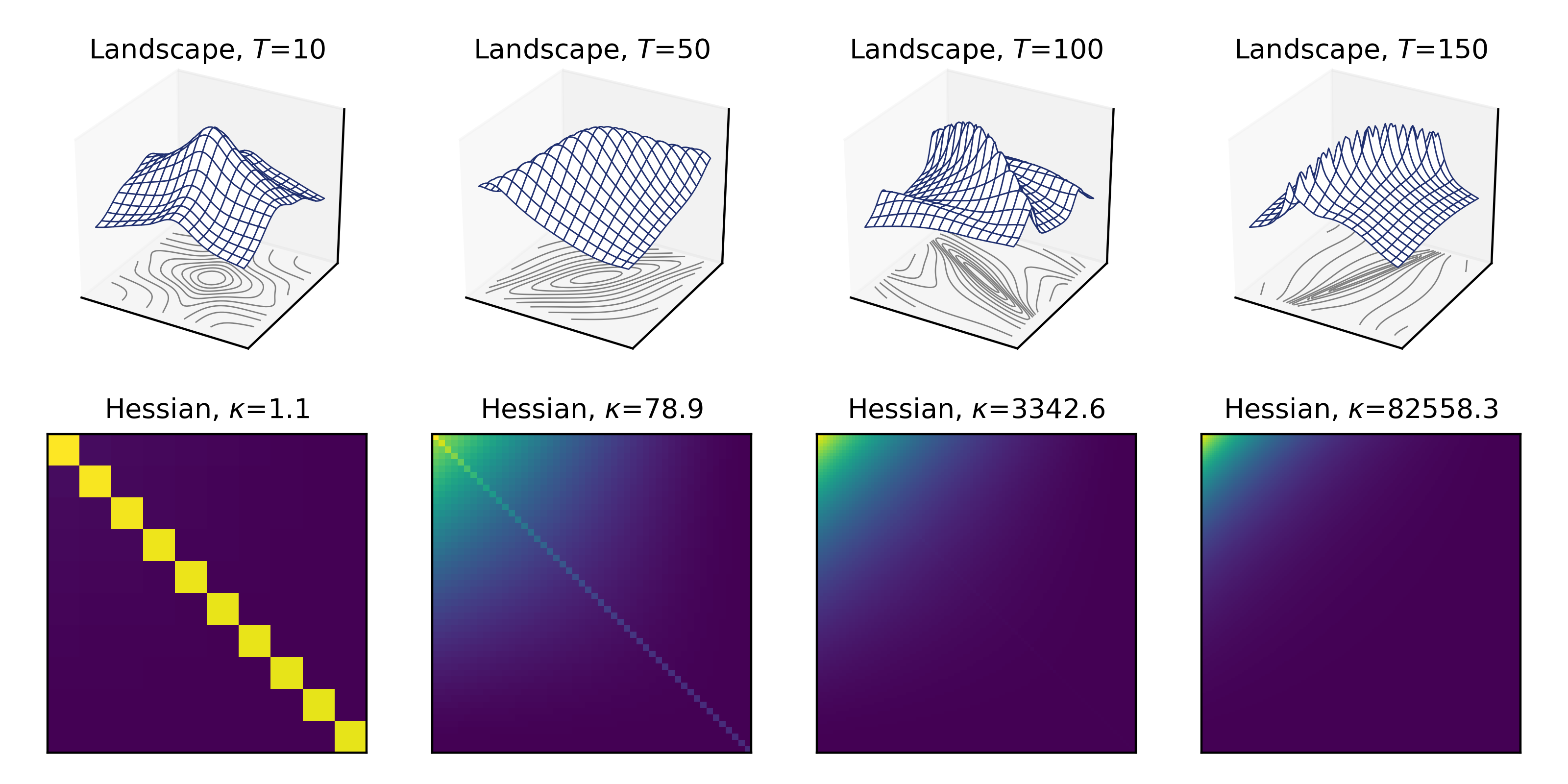}
\caption{Effect of trajectory length $T$ on inverted pendulum trajectory optimization using the reward formulation of Equation \ref{eq:trajreward}. The landscape behaves similarly to the cost minimization in Fig. \ref{fig:sweepT}, except for the additional non-convexity at $T=10$, and overall sharper ridges.}  \label{fig:sweepT_rewards}
\end{figure}

Fig. \ref{fig:sweepT_rewards} shows the inverted pendulum trajectory optimization landscape using the reward function of Equation \ref{eq:trajreward}. Comparing this to Fig. \ref{fig:sweepT}, one notices the following:

\begin{itemize}
\item With large $T$, the landscape structures are essentially the same for both the cost function (Equation \ref{eq:trajcost}) and the reward function (Equation \ref{eq:trajreward}), with an elongated optimum and some local optima. The Hessian and $\kappa(T)$ at the optimal point are the same in both cases. Thus, in principle, the reward function should be as easy or hard to optimize as the cost function. %\textcolor{red}{The above sentence is only true if the problem has the same Hessian everywhere (like QP). But here the Hessian is different across the domain and we only look at the Hessian at the optimum point. I would leave this sentence out.}
\item On the other hand, at $T=10$, the reward landscape shows some additional non-convexity. This suggests that in practice, the reward function may cause a performance hit in optimization. Section \ref{sec:optcompare} provides evidence of this. %\textcolor{red}{It seems to me that the ridges cause multi-modality more so than ill-conditionality or non-convexity (those two issues also exist with cost function). }
\end{itemize}

%Also, even though the $\kappa$ values indicate local similarity, the ridges around the optimum appears overall slightly sharper and more ill-conditioned.

The main difference of the functions is that the sum of exponentiated costs is more tolerant to temporary deviations. A sum-of-squares cost function heavily penalizes the cost being large in even a single timestep, whereas the exponentiation clamps the rewards in the range $[0,1]$. In principle, an agent could exploit the reward formulation by only focusing on some reward terms. We interpret the ridges in Fig. \ref{fig:sweepT_rewards} ($T=10$) as manifestations of this.

In Section \ref{sec:optcompare}, we shall see that the reward function can indeed result in worse optimization performance. However, the next section discusses how using rewards instead of costs can be more desirable when combined with the technique of early termination.

\section{Effect of early termination} \label{sec:termination}
\begin{figure*}[ht]
\centering
\includegraphics[width=7.0in]{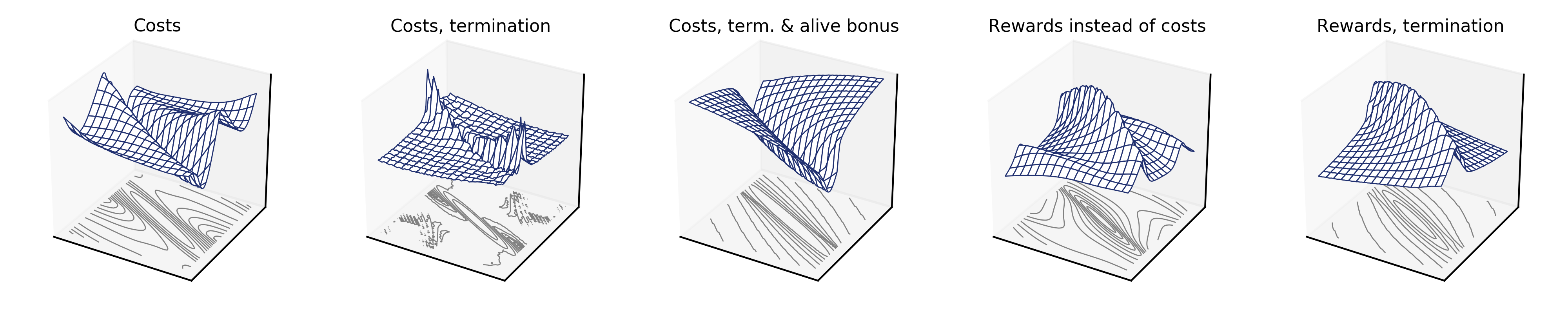}
\caption{Effect of early termination on inverted pendulum trajectory optimization landscape. Termination without an alive bonus increases multimodality of cost minimization, but makes reward maximization more convex.}\label{fig:termination}
\end{figure*}

\begin{figure*}[ht]
\centering
\includegraphics[width=7.0in]{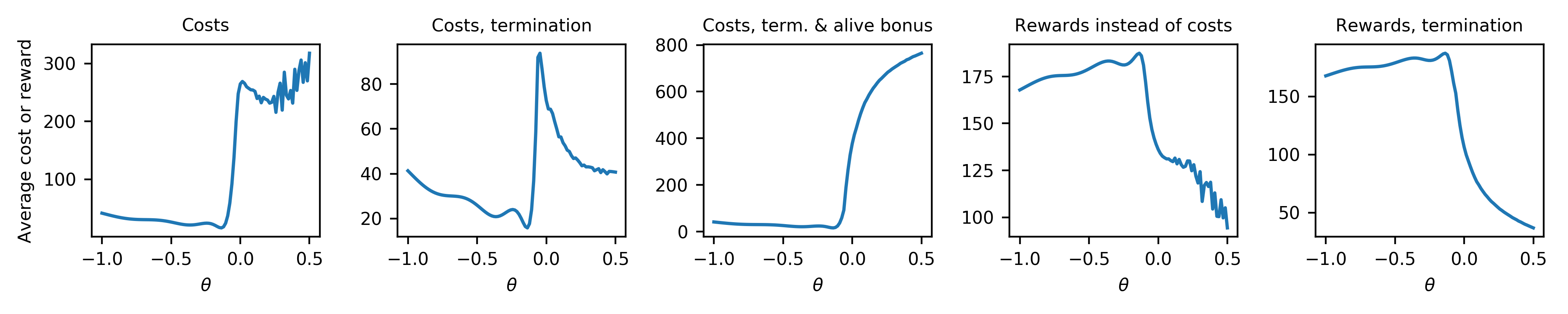}
\caption{Effect of termination on policy optimization. In cost minimization, termination creates a local minimum at $\theta=0.5$, which drives the pendulum to termination to avoid accumulating more costs. Termination removes local optima when combined with an alive bonus or using rewards instead of costs.}\label{fig:termination_policy}
\end{figure*}

Standard continuous control policy optimization benchmark tasks \cite{brockman2016openai,tassa2018deepmind} utilize early termination of simulated episodes; this means that if the agent deviates from some desired region of the state space, e.g. a bipedal agent loses balance, the state is considered a terminal one; the agent stops receiving any rewards, and is reset to an initial state. Typically, termination greatly speeds up policy optimization, e.g. \cite{peng2018deepmimic}. As far as we know, early termination has not been utilized in trajectory optimization, but there are no obstacles for this. %When using termination in Monte Carlo optimization where one samples action sequences and simulates the corresponding state trajectories, one can simply stop the simulation if the termination condition is satisfied, leaving the trajectory cost or reward unaffected by later states and actions. In should be possible to use termination with gradient-based trajectory optimization as well, if one only evaluates and applies the gradient up to the current termination time step. After taking the gradient step, one should re-evaluate the forward dynamics to compute whether and at which timestep the updated trajectory is to be terminated.

Figures \ref{fig:termination} and \ref{fig:termination_policy} show the effect of terminating trajectories and episodes if $|\alpha| > 2.0$, i.e. if the pendulum deviates significantly from the desired upright pose. From the figures, one can gain two key insights:

\begin{itemize}
\item \textit{Termination can greatly improve landscape convexity by removing false optima.} The pendulum cannot receive rewards from the local optimum of hanging downwards if termination prevents it from experiencing the corresponding states. 

\item \textit{If the rewards are not strictly non-negative, new false optima are introduced to the landscape}. Naturally, if the agent is experiencing costs or negative rewards, a good strategy may be to terminate episodes as early as possible, which is what happens at the $\theta=0.5$ local optimum in Fig. \ref{fig:termination_policy} (second subfigure from left, when termination is used with cost minimization). The problem can be mitigated by adding a termination penalty to the cost of the terminal simulation step, or a so-called alive bonus for all non-terminal states. This yields a clearly more convex landscape. 
\end{itemize}

%Although an alive bonus is used in many papers and benchmark tasks \cite{rajeswaran2016epopt, brockman2016openai, yu2018learning}, it is in practice easy to forget about; it is one of the most common mistakes we have identified when supervising students. Thus, our novel visual evidence of the danger of combining termination and negative rewards should be valuable. %\textcolor{red}{TODO: run reinforcement learning on MuJoCo and Unity envs that terminate, check how many give negative rewards} 
% NOTE: is this true? Cannot find examples
%\textcolor{red}{see below paragraph as replacement of the above paragraph?}

Although an alive bonus is used in many papers and benchmark tasks \cite{rajeswaran2016epopt, brockman2016openai, yu2018learning}, we feel the danger of combining termination and negative rewards is not emphasized enough in previous literature. Our visualization highlights that the early termination can be a double-edged sword---it can be harmful when the reward function consists of a mixture of positive and negative terms that are not well-balanced. Our visualization technique suggests that a good default strategy may be to use a combination of termination and rewards instead of costs. This way, one does not need to fine-tune the termination penalty or alive bonus. Furthermore, converting costs to rewards through exponentiation limits the results to the range $[0,1]$, which is beneficial for most RL algorithms that feature some form of value function learning with neural networks. On the other hand, as shown in the next section, it can result in slower convergence.

\section{Do visualizations predict optimization performance?}\label{sec:optcompare}
\begin{figure}[th]
\centering
\includegraphics[width=\linewidth]{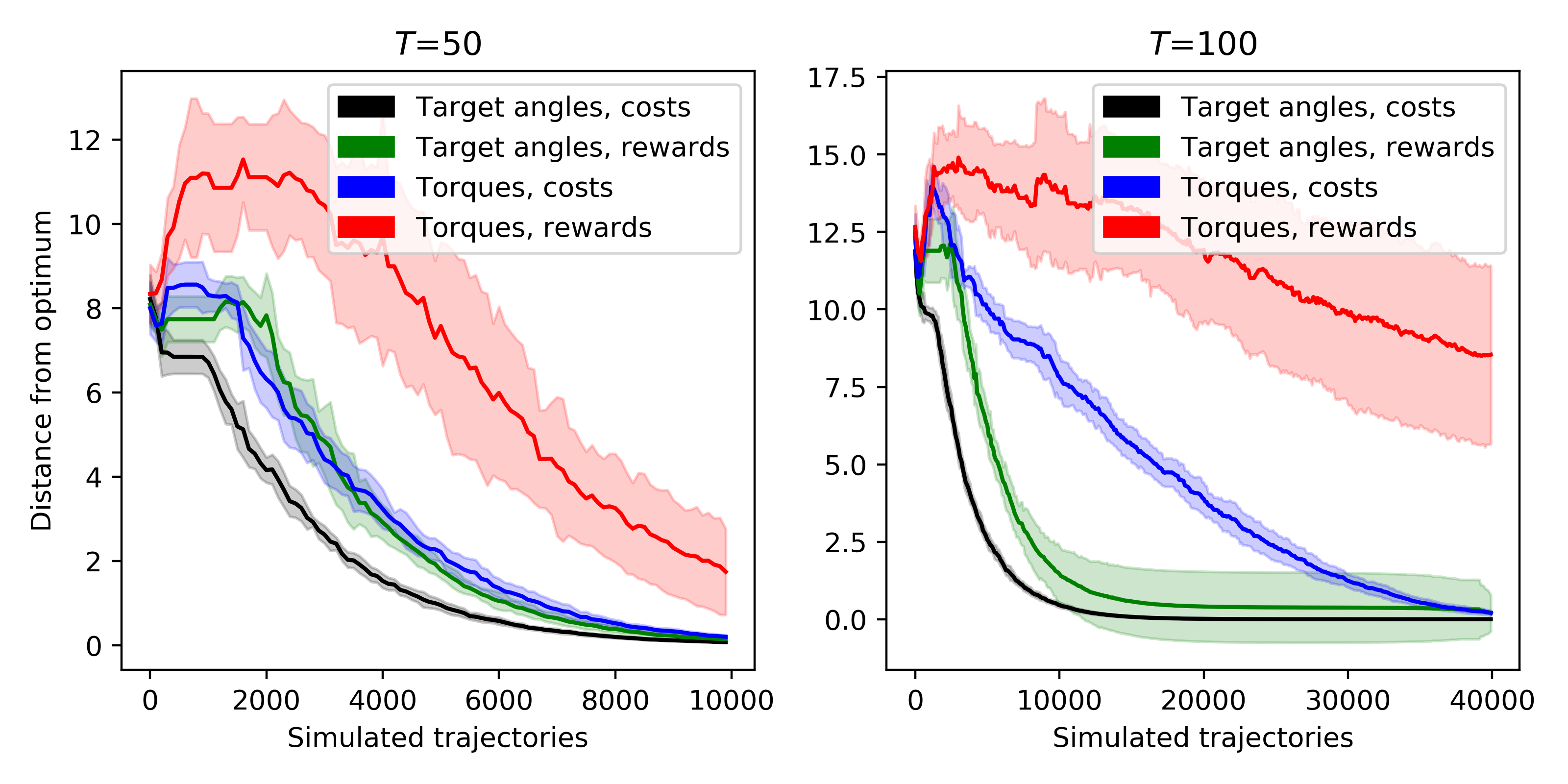}
\caption{Inverted pendulum trajectory optimization results, plotted as the mean of 10 runs of CMA-ES. We compare both cost minimization and reward maximization with actions parameterized both as torques and target angles. As predicted by our visualizations, the angle parameterization scales better for large $T$, and the reward maximization is less efficient.} \label{fig:optCompare}
\end{figure}

Fig. \ref{fig:optCompare} tests how well the visualizations of the previous sections predict actual optimization performance in the inverted pendulum trajectory optimization. The figure compares quadratic costs to exponentiated rewards, and actions parameterized as torques to target angles implemented using a P-controller as in Section \ref{sec:actionspace}. To keep the figure readable, we did not include curves with and without termination; this does not make a big difference in the simple pendulum problem, and termination is further investigated in the next section. All the optimizations were performed using CMA-ES \cite{hansen2001completely,hansen2016cma}, which is common in animation research, and is known to perform well with multimodal optimization tasks. A population size of 100 was used. To allow comparing both costs and rewards, progress is graphed as the Euclidian distance from the true optimum.

The results indicate that parameterizing actions as target angles is considerably more efficient, as predicted by the visualizations of Section \ref{sec:actionspace}. The exponential transform of costs to rewards degrades performance, in line with the observations of Section \ref{sec:rewards}. %The random 2D landscape visualizations appear more informative than $\kappa$ of the Hessian, which would predict that using torques is better than target angles with both $T=50$ and $T=100$.
%\item In line with Section \ref{sec:termination}, combining costs and termination is detrimental, in particular with a longer planning horizon.
 
%\item Early termination does not have an effect, but this is not surprising considering that the inverted pendulum landscape has only mild multimodality that CMA-ES has no problems with even if termination is not used to remove local optima. %In the case of the bipedal agent of the next section, termination does help.

%Appendix \ref{sec:theory} provides additional mathematical analysis of the reliability of the visualizations.

%shows how our visualizations are visualization is indicative of easier optimization; on average, optimizing target angles converges much faster. We perform the optimization both with CMA-ES \cite{hansen2001completely,hansen2016cma} and Adam \cite{kingma2014adam}. For the latter, the pendulum dynamics were we implemented using Tensorflow \cite{abadi2016tensorflow}, allowing gradient-based optimization usingbackpropagation through time. 

\section{Generalizability to more complex agents}\label{sec:biped}

\begin{figure}[t]
  \centering
  \includegraphics[width=\linewidth]{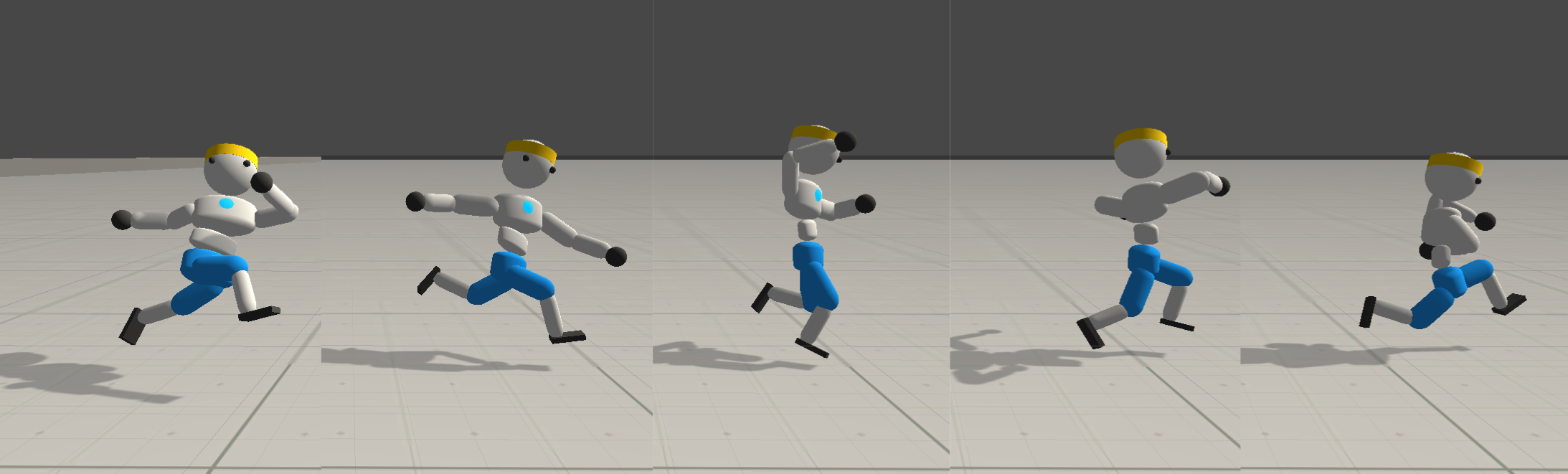}
  \caption{3D humanoid locomotion test from Unity Machine Learning Agents framework.} \label{fig:humanoid}
\end{figure}

This section tests the generalizability and usefulness of the visualization approach with a more complex agent, using both policy optimization and trajectory optimization. We use the 3D humanoid locomotion test from the Unity Machine Learning Agents framework \cite{juliani2018unity} shown in Fig. \ref{fig:humanoid}. The test is designed for policy optimization; we modify it to also support trajectory optimization by starting each episode/trajectory from a fixed initial state, without randomization. We have also modified the code and environment to be fully deterministic, i.e., non-reproducible simulation is not corrupting the landscape visualizations. 

The action space is 39-dimensional, with actions defined as both joint target angles, as well as maximum torques that the joint motors are allowed to use for reaching their targets. We optimize with planning horizons of 1, 3, 5, and 10 seconds, resulting in 585, 1755, 2925, and 5850 optimized variables. The actions define target angles and maximum torques for the 16 joints of the humanoid. As the global optima are not known, we visualize the landscapes around the optima found in each test, following Li et al. \cite{li2018visualizing}. We use a simulation timestep of $1/75$ seconds, with control actions repeated for 5 timesteps, i.e., taking 15 actions per second. 

As detailed below in Sections \ref{sec:biped_traj}-\ref{sec:scalability}, the results support our earlier findings:

\begin{itemize}
\item 2D visualization slices reveal that trajectory optimization is highly multimodal, with optima that become narrower as the length of the planning horizon increases.
\item As hypothesized in Section \ref{sec:splines}, utilizing a spline parameterization results in a more well-behaving landscape. % (Fig. \ref{fig:humanoid_spline}.
\item Termination based on agent state reduces local optima in both trajectory and policy optimization.% (figures \ref{fig:humanoid_termination}-\ref{fig:humanoid_policy}).
\item Policy optimization with neural network policies scales better for long planning horizons, with the landscape remaining essentially unchanged as the planning horizon grows. Notably, \textit{policy optimization was more efficient than optimizing a single long trajectory}, even though our policy network has over 2M parameters to optimize, i.e., orders of magnitude more. 
\end{itemize}

\subsection{Trajectory Optimization}\label{sec:biped_traj}
For trajectory optimization we use the recent highly scalable CMA-ES variant called LM-MA-ES \cite{loshchilov2018large}---as the physics simulator is not differentiable, gradient-based methods are not applicable. We visualize the landscapes around the optimum found by LM-MA-ES. Similar to CMA-ES, LM-MA-ES is a quasi-parameter-free method, typically only requiring adjustments to the iteration sampling budget (population size): this is increased from the recommended value for more difficult problems. The recommended budget---a logarithmic function of the number of variables---did not produce robust results. Instead, we used a 10 times larger budget in all the optimization runs.

We first tested trajectory optimization on action sequences of up to 5 seconds, i.e., up to 2925 optimized variables. This turned out to be a difficult task, resulting in somewhat unstable gaits even after hours of CPU time.%Despite the reported high efficiency of LM-MA-ES with high-dimensional optimization tasks, we were unable to find a stable gait with trajectory optimization.

\begin{figure}[t]
  \centering
  \includegraphics[width=\linewidth]{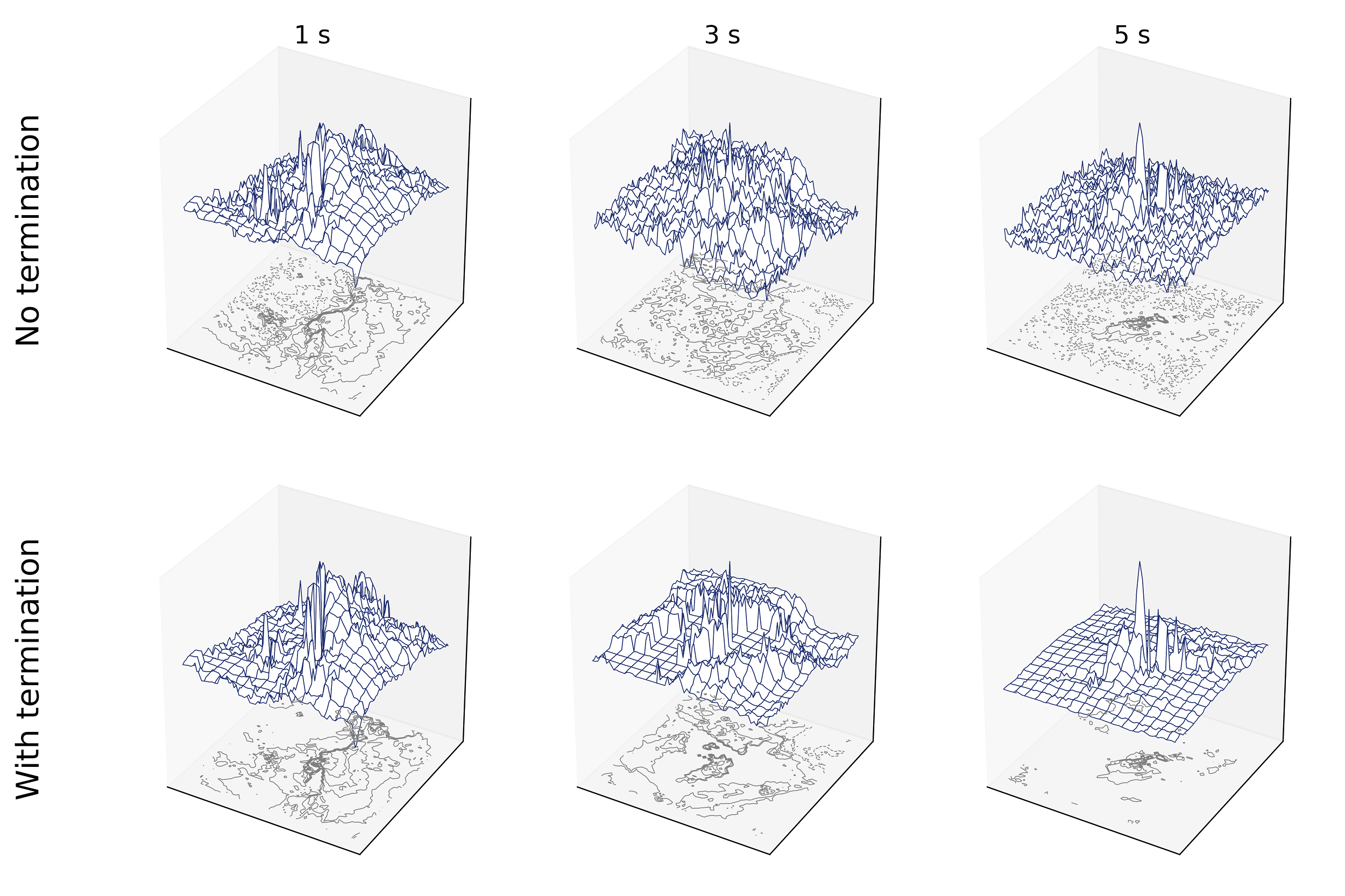}
  \caption{Trajectory optimization landscapes of humanoid locomotion with different planning horizons. Termination removes local optima, producing a smoother landscape.} \label{fig:humanoid_termination}
\end{figure}

Here, 2D slice visualizations provided useful diagnostic information, revealing the difficult multimodality of the optimization problem. Fig. \ref{fig:humanoid_termination} shows the landscapes around the found local optimum for each tested planning horizon. Although there is a clear optimum in the center, the rest of the landscape is noisy and ill-conditioned. This is exacerbated with the longer planning horizons. 

Fig. \ref{fig:humanoid_termination} also shows that termination helps remove local optima, and produces a smoother landscape. In the humanoid locomotion test, termination takes place when a body part other than the feet touches the ground. 

\subsection{Trajectory Optimization with Splines} 
\begin{figure}[t]
  \centering
  \includegraphics[width=\linewidth]{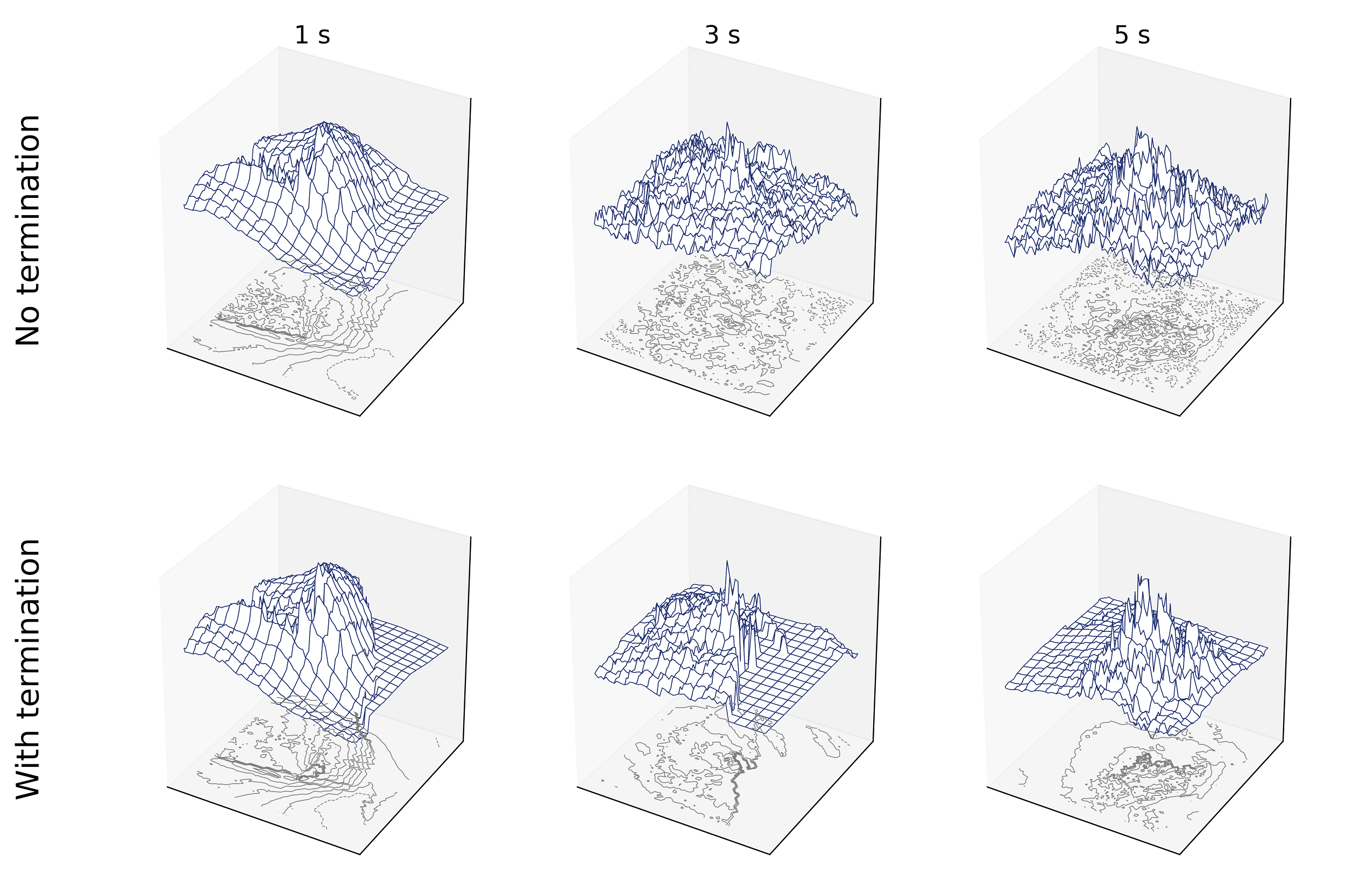}
  \caption{Spline-parameterized trajectory optimization landscapes of humanoid locomotion. The landscapes exhibit less noise than the trajectory optimization landscapes of Fig. \ref{fig:humanoid_termination}\label{fig:humanoid_spline}} 
\end{figure}

In an effort to better handle long planning horizons, we tested trajectory optimization with spline-based parameterization. Instead of setting target angles and maximum torques for joints once every 5 timesteps, we interpolated the actions for each timestep using Catmull-Rom splines. The control points of the splines were optimized using LM-MA-ES, akin to the method presented by Hämäläinen et al. \cite{Hamalainen2014} for online optimization. Fig. \ref{fig:humanoid_spline} shows how this slightly improves the landscapes, with reduced noise and gentler slopes towards the optimum. %Notably, the 5 second horizon in particular shows improvement over the earlier landcapes in Fig. \ref{fig:humanoid_termination}, having a more continuous slope towards the optimum.  

\begin{figure}[t]
  \centering
  \includegraphics[width=\linewidth]{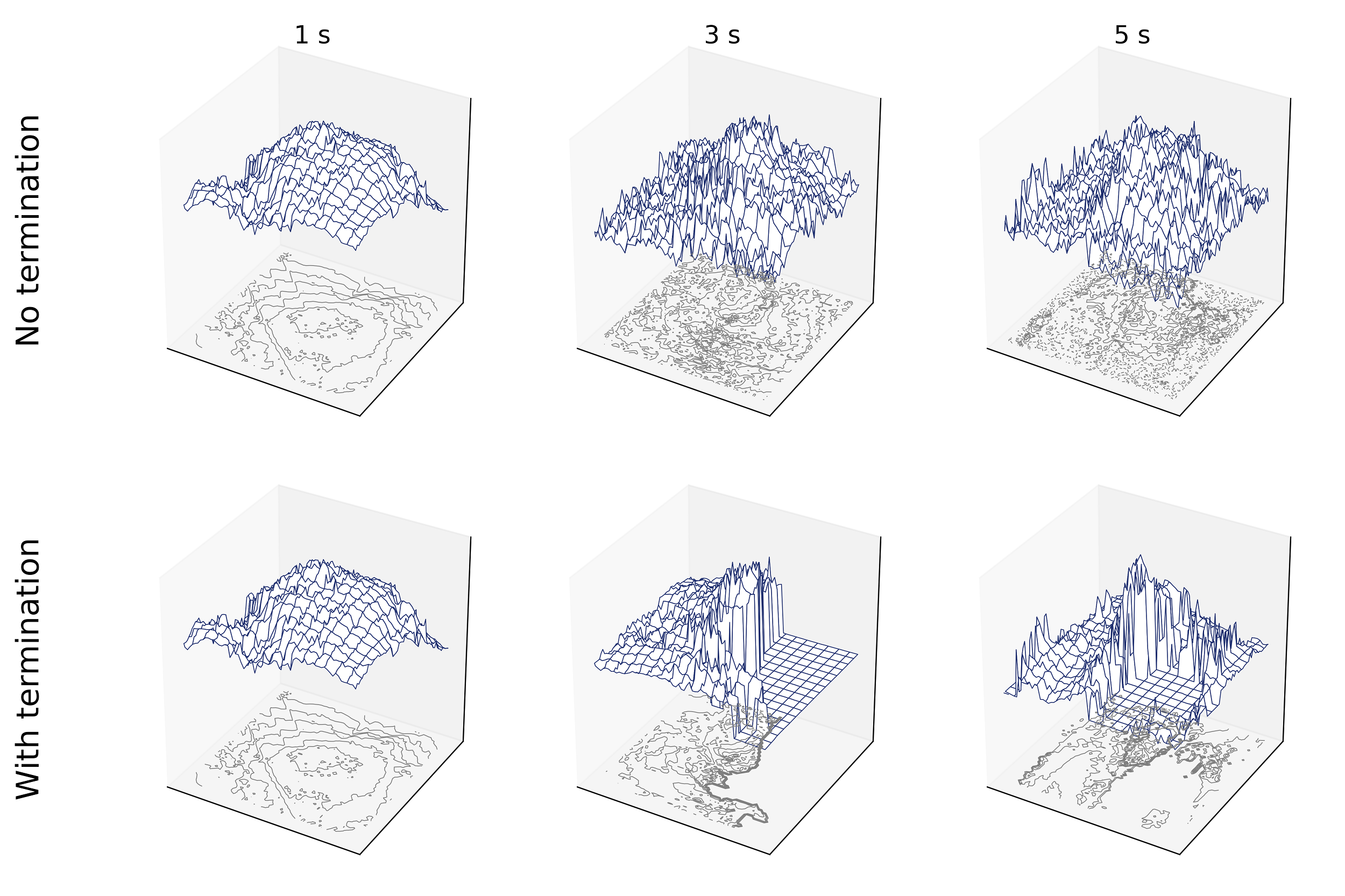}
  \caption{Non-spline trajectory optimization landscapes around action sequences resulting from evaluating the optimal splines of Fig. \ref{fig:humanoid_spline}. The landscapes become noisier and introduce more elongated ridges and valleys.} \label{fig:humanoid_per_time_step}
\end{figure}

To provide a more direct comparison to optimizing action sequences without splines, Fig. \ref{fig:humanoid_per_time_step} shows the non-spline trajectory optimization landscape around the action sequences resulting from the optimized splines of Fig. \ref{fig:humanoid_spline}. Without the spline parametrization, the 3s and 5s landscapes degenerate, becoming noisier and more ill-conditioned. %When comparing the 5 second non-spline action trajectory, there is a dramatic increase in peaks and valleys. 

Although trajectory optimization with splines results in cleaner landscapes and qualitatively better and smoother movement than non-spline trajectory optimization, finding good long trajectories required hours of CPU time. This motivated us to also test policy optimization, as explained below.  % where the optimized variables do not necessarily have as strong dependencies. %While this supports the usage of spline-parametrization, many of the problems with the optimization task are retained. The landscapes remain noisy and non-convex.

\subsection{Policy Optimization} \label{sec:biped_PPO}
We trained a neural network policy for solving the humanoid locomotion task using Proximal Policy Optimization (PPO) \cite{schulman2017proximal} and different planning horizons. PPO utilizes episodic experience collection, i.e., the agent is started from some initial state, and explores states and actions until a terminal state, or maximum episode length, is reached. We used the planning horizon as the episode length.

\begin{figure}[t]
  \centering
  \includegraphics[width=\linewidth]{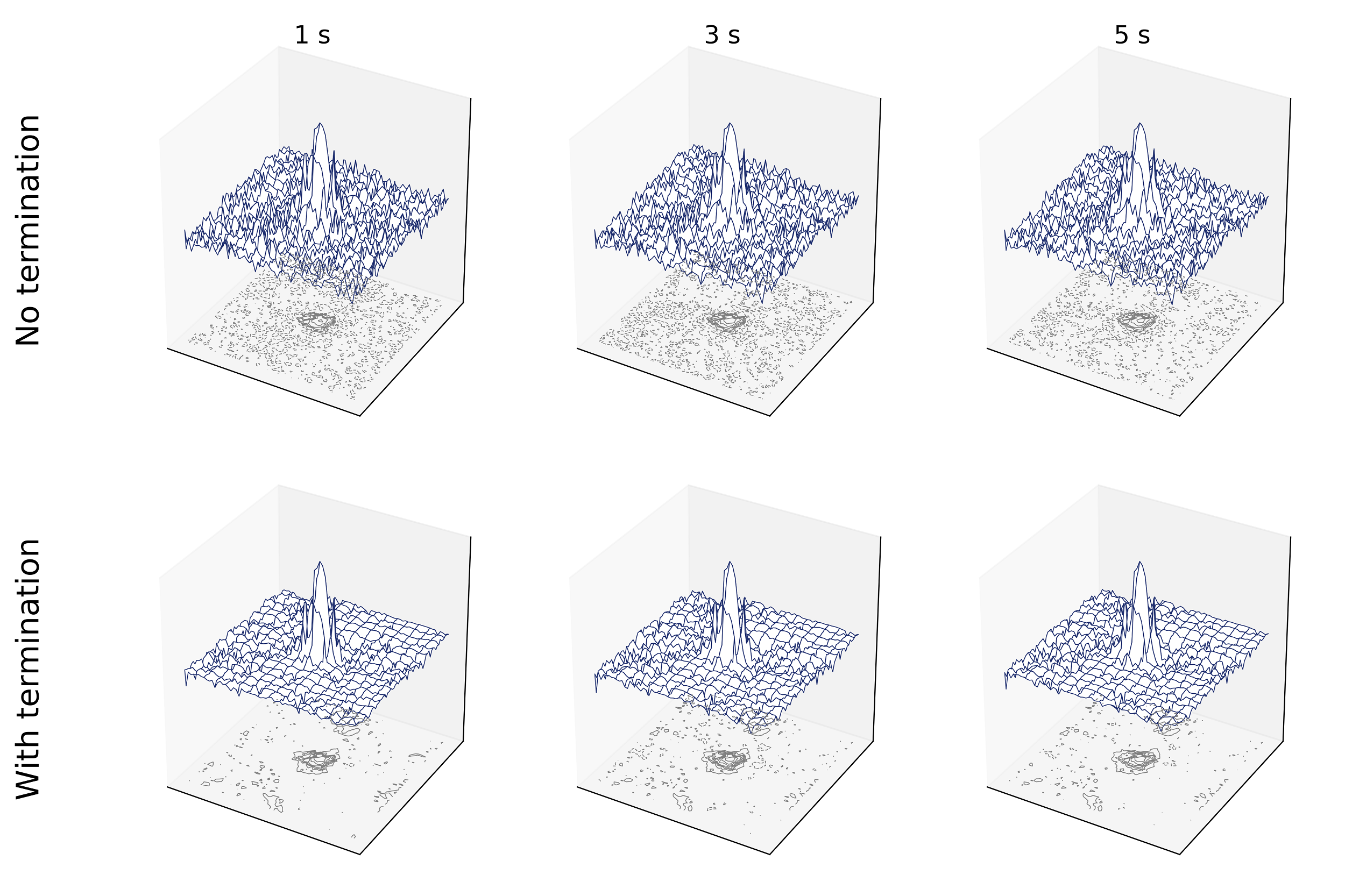}
  \caption{Policy optimization landscapes of humanoid locomotion. As opposed to trajectory optimization, the task scales well with increasing planning horizon. Termination removes local optima.} \label{fig:humanoid_policy}
\end{figure}

The resulting landscapes, shown in Fig. \ref{fig:humanoid_policy}, show a significant improvement in convexity, sphericity, and unimodality. The landscape also remains essentially unchanged with a longer planning horizon.

\begin{figure}[t]
  \centering
    \includegraphics[width=\linewidth]{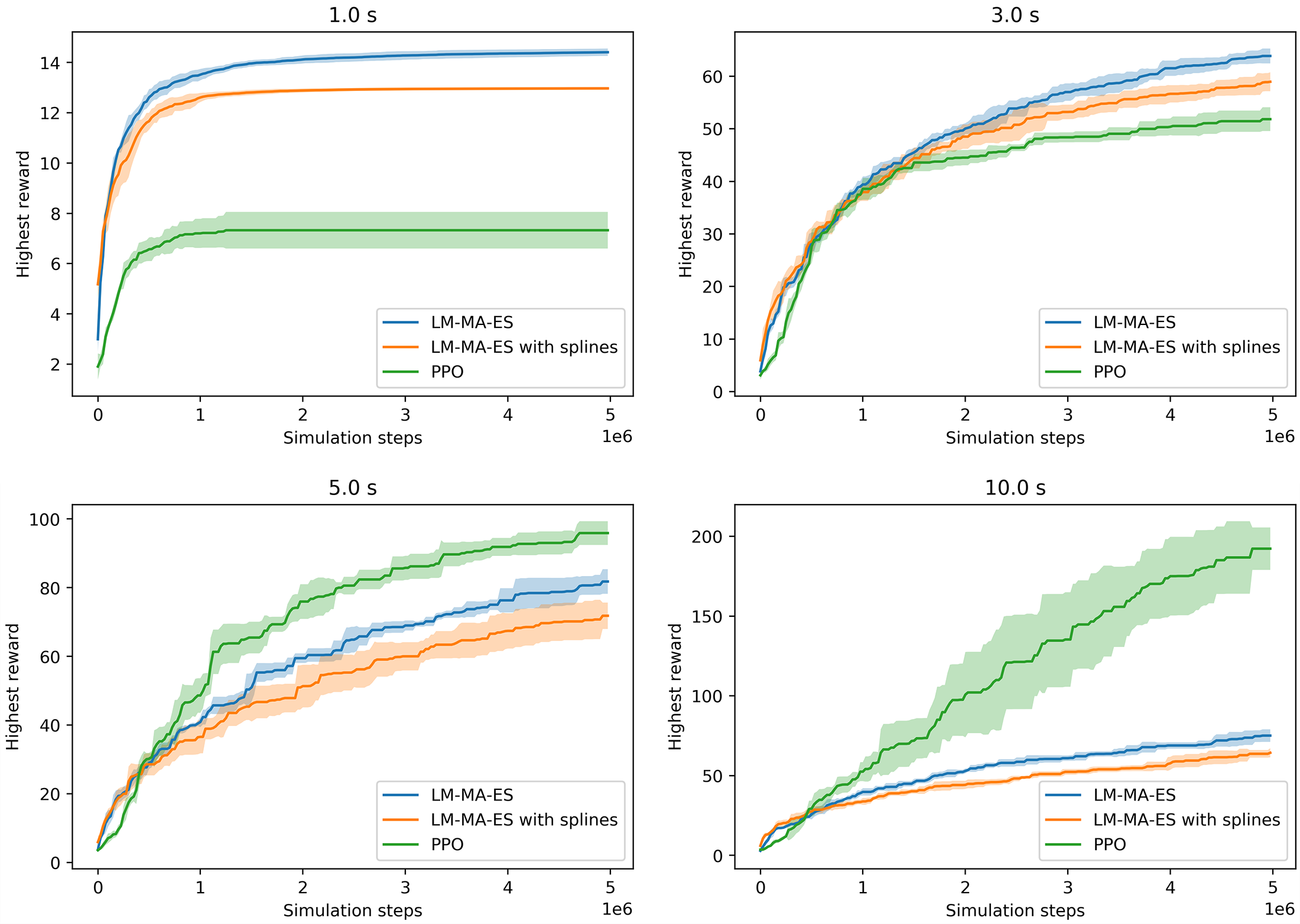}
%  
%  \begin{subfigure}[b]{0.49\linewidth}
%    \includegraphics[width=\linewidth]{images/Convergence_1.png}
%  \end{subfigure}
%  \begin{subfigure}[b]{0.49\linewidth}
%    \includegraphics[width=\linewidth]{images/Convergence_3.png}
%  \end{subfigure}
%  \begin{subfigure}[b]{0.490\linewidth}
%    \includegraphics[width=\linewidth]{images/Convergence_5.png}
%  \end{subfigure}
%  \begin{subfigure}[b]{0.49\linewidth}
%    \includegraphics[width=\linewidth]{images/Convergence_10.png}
%  \end{subfigure}
  \caption{Best trajectory or episode reward as a function of timesteps simulated during optimization, with different planning horizons and optimization approaches. The graphs show the mean and standard deviation of 5 independent optimization runs. Policy optimization with PPO scales significantly better for long planning horizons. \label{fig:humanoid_comparison}} 
\end{figure}

\subsection{Scalability of Trajectory and Policy Optimization}\label{sec:scalability}
The policy optimization visualizations suggest that as trajectory lengths increase, policy optimization should be more efficient than trajectory optimization. Fig. \ref{fig:humanoid_comparison} provides evidence supporting this hypothesis. It compares  optimization progress with different planning horizons and the three optimization strategies: LM-MA-ES, LM-MA-ES with splines, and PPO. With a trajectory length of 5 seconds, PPO already shows improved performance over the other methods, and the advantage is dramatically larger with 10 second trajectories. However, this comes with a caveat: with a neural network policy, there is more overhead per simulation step, as each optimization iteration requires training the policy and value networks using multiple minibatch gradient updates. Optimizing for 5 million timesteps with PPO was approximately 4 times slower than with LM-MA-ES in our tests, when measured in wall-clock time, using a single 4-core computer. 

Interestingly, although the spline landscapes look slightly better in Fig. \ref{fig:humanoid_spline}, spline trajectory optimization results in slightly lower rewards with a given simulation budget. A plausible explanation for this is that the Unity locomotion test has a very simple reward function that trajectory optimization can exploit with unnatural and jerky movements, as shown on the supplemental video\footnote{\url{https://youtu.be/5v_lsGCahSI}} at 03:01. From a pure reward maximization perspective, a spline parameterization that enforces a degree of smoothness may not be ideal and one may expect more gains if the reward function favors smooth movement. In general, best results are achieved when the action parameterization induces a useful prior for the optimization task.

\begin{figure}[t]
    \includegraphics[width=\linewidth]{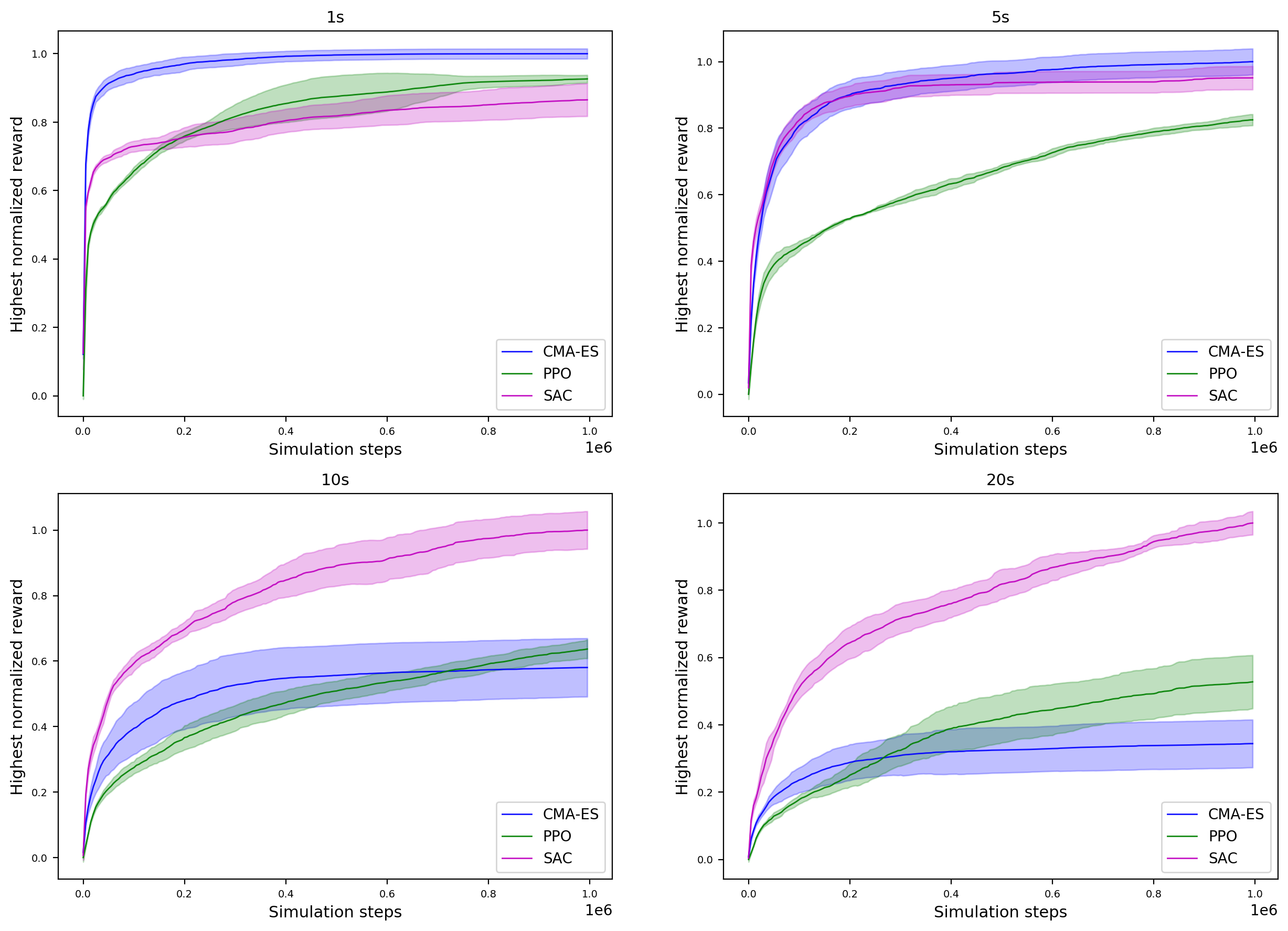}
  \caption{Replicating the result of Fig. \ref{fig:humanoid_comparison} in 4 OpenAI Gym MuJoCo tasks (Walker2d-v2, Hopper-v2, HalfCheetah-v2, Humanoid-v2), using 3 optimization methods (CMA-ES, PPO, and SAC), planning horizons 1s, 5s, 10s, and 20s, and 10 independent optimization runs per method and task. Each learning curve aggregates the results from all 4 tasks, using normalized episode/trajectory rewards. \label{fig:mujoco-results}} 
\end{figure}

Fig. \ref{fig:mujoco-results} replicates the result of Fig. \ref{fig:humanoid_comparison} with other agents and optimization methods, providing additional evidence that policy optimization scales better to long planning horizons. To generate the figure, we conducted trajectory and policy optimizations using 4 common OpenAI Gym \cite{brockman2016openai} MuJoCo agents and locomotion tasks (or ``environments'' in the Gym lingo): 2D monopedal hopper (Hopper-v2), 2D bipedal walker (Walker2d-v2), 2D half quadruped (HalfCheetah-v2), and 3D humanoid (Humanoid-v2). In policy optimization, we tested both PPO and Soft Actor-Critic (SAC) \cite{haarnoja2018soft}, a more recent method that is growing in popularity. We used the Stable Baselines \cite{stable-baselines} PPO and SAC implementations with their default settings. In trajectory optimization, we used CMA-ES instead of LM-MA-ES, as it was easily available for the Python-based Gym framework. For each task and optimization method, we performed 10 independent training runs with different random seeds. To allow aggregating the convergence curves of all tasks in a single plot, we normalized the episode/trajectory rewards of each task over all optimization runs and methods to the range [0,1]. We used the default MuJoCo reward functions and episode termination, but removed the initial state randomization to allow direct comparison of trajectory and policy optimization, similar to the Unity humanoid tests above. We used a control frequency of 20Hz for all the tasks.

%The mathematical analysis was originally in appendix, now included in the body
\input{appendix_in_body}

\section{Conclusion}
We have presented several novel visualizations of continuous control trajectory and policy optimization landscapes, demonstrating the usefulness of the random 2D slice visualization approach of Li et al. \cite{li2018visualizing} in this domain. We have also presented a mathematical analysis of the limitations of the random 2D slice visualizations. 

Our visualizations provide new intuitions of movement optimization problems and explain why common best practices are powerful. The visualization approach can be used as a diagnostic tool for understanding why optimization does not converge, or progresses slowly. Even when a global optimum is not known, as in Section \ref{sec:biped}, it can be useful to plot the landscape around a found optimum: If the optimized movement is not satisfactory or one optimization approach performs worse than another, visualization can provide insights on why this is the case, e.g., due to ill-conditioning or multimodality. %Similar visualization can be used to confirm that a problem modification had a desired effect.

We acknowledge that some of our results---e.g. the efficiency of episode termination---are already known to experienced readers. We do, however, provide novel visual evidence of the underlying reasons; for example, we show how termination based on agent state removes local optima in the space of optimized actions or policy parameters. This contributes to the understanding of movement optimization, and, as representative images are known to increase understanding and recall \cite{carney2002pictorial}, it should also have pedagogical value in educating new researchers and practitioners.

To conclude, the key insights from our work can be summarized as:

\begin{itemize}
\item Random 2D slice visualizations are useful in analyzing high-dimensional movement optimization landscapes, and can predict movement optimization efficiency. % If a random 2D visualization slice shows nonconvexity or ill-conditioning, they are real problems that may need to be addressed. At least in the optimization problems of this paper, the 2D visualizations predict optimization performance better than the condition number of the Hessian. %However, it may sometimes be useful to select the visualization basis vectors such that one focuses on the actions of consecutive timesteps. 
\item The curse of dimensionality hits trajectory optimization hard, as it can become increasingly ill-conditioned with longer planning horizons. Policy optimization scales better in this regard. Perhaps counterintuitively, optimizing a neural network policy can be more efficient than optimizing a single action trajectory with orders of magnitude less parameters.
\item Parameterizing actions as (partial) target states---e.g. target angles that the character's joints are driven towards---is strongly motivated, as opposed to optimizing raw control torques. It can make trajectory optimization more well-conditioned and separable. %Note that our results are optimistic, as the target angle of a pendulum defines the state more completely than the target pose of a humanoid character; the latter leaves the character root unactuated and thus does not remove all dependencies between actions. Regarding future work, it appears crucial to replace P- and PD-controllers with more advanced controllers -- e.g., neural network policies -- that allow reaching a more completely defined target state even for complex characters like humanoids.
\item Combining the two points above, one can explain the power of the common practice of optimizing splines that define time-varying target poses; pose parameterization leads to more well-conditioned optimization, and the spline control points define a shorter sequence of macro actions, which further counteracts ill-conditioning caused by sequence length. However, the smoothness constraints that splines impose on movements may not be ideal for all reward functions.
\item Using early termination appears strongly motivated, as it typically results in a more convex landscape in both trajectory and policy optimization. However, combining termination with costs or negative rewards is dangerous, which our visualizations clearly illustrate. %Both our visualizations and empirical convergence results also suggest that an exponential transform of cost to rewards is not optimal; an alternative is to use costs with an alive bonus that ensures that the total reward is not negative.
\end{itemize}

In our future work, we aim to investigate the parameterization of actions as target states for complex controlled agents with unactuated roots. We hypothesize that this can be implemented for both trajectory and policy optimization, using a general-purpose neural network controller trained for reaching the target states. %, through curiosity-driven exploration of the state space, potentially guided by a corpus of motion capture data.

 %For example, Figures 14 and 15 verify the intuition that contact discontinuities and a long planning horizon might result in a multimodal, discontinuous trajectory optimization landscape that is hard to optimize. %Note that the jaggedness of the landscapes is not due to dynamics noise; the simulation is deterministic.

%Even though some of the insights above may be obvious to an experienced reader, we believe that our visualizations still have at least pedagogical value for teaching control optimization and computer animation.  
%It should be noted that we have only focused on the recently popular case where one has a forward dynamics simulator that may be a black-box one, typically requiring Monte Carlo methods for optimization. In future work, it would be interesting to visualize and compare these landscapes with earlier research on spacetime optimization \cite{witkinkass,mordatchcontactinvariant}, where both the dynamics and the control are combined into one big optimization problem, that has shown to be tractable without Monte Carlo methods.

\section*{Acknowledgements}
This research has been supported by Academy of Finland grant 299358.

% Bibliography
\bibliographystyle{IEEEtran}
\bibliography{perttu}

%% file: appendix_in_body.tex
%!TEX root = main.tex

%=======================================================================
\section{Properties and limitations of random 2D slice visualizations}\label{sec:theory}
So far, we have provided many examples of random 2D slice visualizations of high-dimensional objective functions. We have also demonstrated that such visualizations have predictive power regarding the difficulty of optimization. However, as information is obviously lost in only evaluating the objective along a 2D slice, this section provides further analysis of the limitations of the approach. To allow investigating how the mathematical properties of objective functions manifest in the visualized slices, this section focuses on simple test functions with closed-form expressions. Such simple expressions are not available for real-life movement optimization objectives that depend on complex simulated dynamics, typically implemented using a black-box physics simulator.

\subsection{2D Visualizations are Optimistic About Ill-conditioning}
%As elaborated below, the number of high and low eigenvalues of the Hessian greatly affects visualization in addition to the condition number, i.e., ratio of highest and lowest eigenvalues.
Consider the following cost function:% with intrinsic dimensionality $k$:  
\begin{eqnarray}
f(\mathbf{x})&=&\sum_{i=1}^{k} x_i^2 + \epsilon \sum_{i=k+1}^{d}  x_i^2\\
&=& ||\mathbf{x}_{:k}||^2 + \epsilon ||\mathbf{x}_{k:}||^2, \label{eq:kd}
\end{eqnarray}
where $\epsilon$ is a small constant, i.e., $f(\mathbf{x})$ mostly depends only on the first $k$ optimized variables. $\mathbf{x}_{:k}$ denotes the projection of $\mathbf{x}$ into the subspace of the first $k$ dimensions. $\mathbf{x}_{k:}$ denotes the projection into the remaining dimensions. The Hessian of $f(\mathbf{x})$ is diagonal, containing the curvatures along the unit vectors as the diagonal elements. Curvature along the first $k$ unit vectors equals $2$ and curvature along the rest of the dimensions is $2\epsilon$. Thus, if $k \neq 0$ and $k \neq d$, the condition number $\kappa = 1/\epsilon$. 

\begin{figure}[h]
\centering
\includegraphics[width=\linewidth]{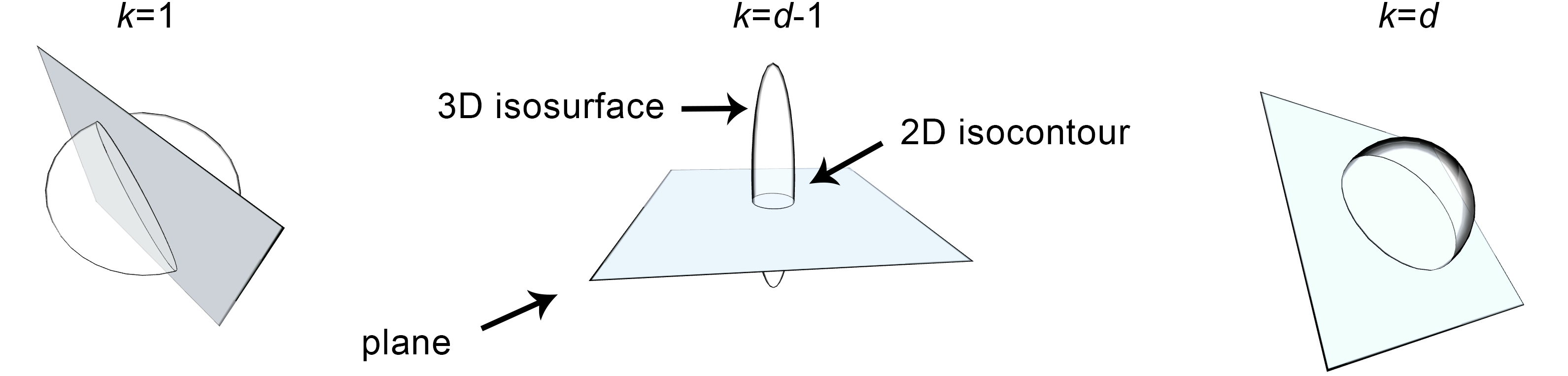}
\caption{The isosurfaces, random visualization planes, and 2D visualization isocontours for Equation \ref{eq:kd} in the case of $d=3$.}\label{fig:isosurfaces}
\end{figure}

Geometrically, the isosurfaces of $f(\mathbf{x})$ are $n$-spheres elongated by a factor of $1/\sqrt{\epsilon}$ along the last $d-k$ dimensions. The visualized 2D isocontours correspond to the intersections of the isosurfaces with the visualization plane. This is illustrated in Fig. \ref{fig:isosurfaces} and on the supplemental video for $d=3$.

%Investigating Fig. \ref{fig:isosurfaces} reveals a basic property of the 2D visualizations:

%\medskip
%\textbf{Proposition 1} \textit{With the convex quadratic objective of Equation \ref{eq:kd}, the elongation of the 2D isocontours is less than or equal to the true elongation of the isosurfaces. In other words, $\kappa_{2D} \le \kappa$.}

%\textit{Proof.} The proposition follows from the isocontours corresponding to planar intersections of the isosurfaces. First, as illustrated in the middle of Fig. \ref{fig:isosurfaces}, it is possible to rotate the visualization plane such that the isocontours display less elongation. Second, the isocontours cannot display more than the real elongation; the visualized elongation is at maximum when one plane basis vector aligns with a direction of high elongation (vertical axis in the middle of Fig. \ref{fig:isosurfaces}), and the other basis vector aligns with a direction of low elongation, in which case the isocontours display the correct $\kappa = 1/\epsilon$. $\qed$

%\textcolor{red}{I propose to not call it Proposition because we do not have a formal proof here. Suggested change:}

Investigating Fig. \ref{fig:isosurfaces} reveals a basic property of the 2D visualizations: with the convex quadratic objective of Equation \ref{eq:kd}, the elongation of the 2D isocontours is less than or equal to the true elongation of the isosurfaces. \emph{In other words, $\kappa_{2D} \le \kappa$.}

This property follows from the isocontours corresponding to planar intersections of the isosurfaces. First, as illustrated in the middle of Fig. \ref{fig:isosurfaces}, it is possible to rotate the visualization plane such that the isocontours display less elongation. Second, the isocontours cannot display more than the real elongation; the visualized elongation is at maximum when one plane basis vector aligns with a direction of high elongation (vertical axis in the middle of Fig. \ref{fig:isosurfaces}), and the other basis vector aligns with a direction of low elongation, in which case the isocontours display the correct $\kappa_{2D} = \kappa = 1/\epsilon$.

%illustrates how even if a random 2D visualization shows a well-conditioned objective -- e.g., in the middle of the figure -- the reality may be worse. On the other hand, it is possible to rotate the visualization plane such that it aligns with both the maximal and minimal axes of elongation and thus displays the true ill-conditioning. These observations can be stated as a proposition
%The visualized elongation is at maximum when the plane aligns with the dimensions where the isosurface is the widest and narrowest,

%The case $k=d$ in Fig. \ref{fig:isosurfaces} is simple and can be formally stated as:

%\textit{Proof.} The isosurfaces of a spherical function are $n$-spheres. The isocontours of the visualization are intersections of the visualization plane and the $n$-spheres, i.e., circles. $\qed$

%If $k=d-1$, cost is low only near the last unit vector. If $k=1$, cost is low inside the hyperplane comprising the last $d-1$ dimensions, and cost grows rapidly only when moving along the first unit vector, i.e., the normal of the hyperplane.

%A crucial observation from Fig. \ref{fig:isosurfaces} is that the number of possible visual plane rotations that display ill-conditioning depends on $k$. When $k=1$, the isocontours are mostly highly elongated, except when the visualization plane is aligns with the lens-shaped isosurface. In contrast, with $k=d-1$, the 2D isocontours are mostly not showing the true elongation of the isosurface, except when the visualization plane aligns with the axis of elongation. 

\subsection{2D Visualizations Show Ill-conditioning More Accurately With Low Intrinsic Dimensionality}
\begin{figure}[t]
\centering
\includegraphics[width=\linewidth]{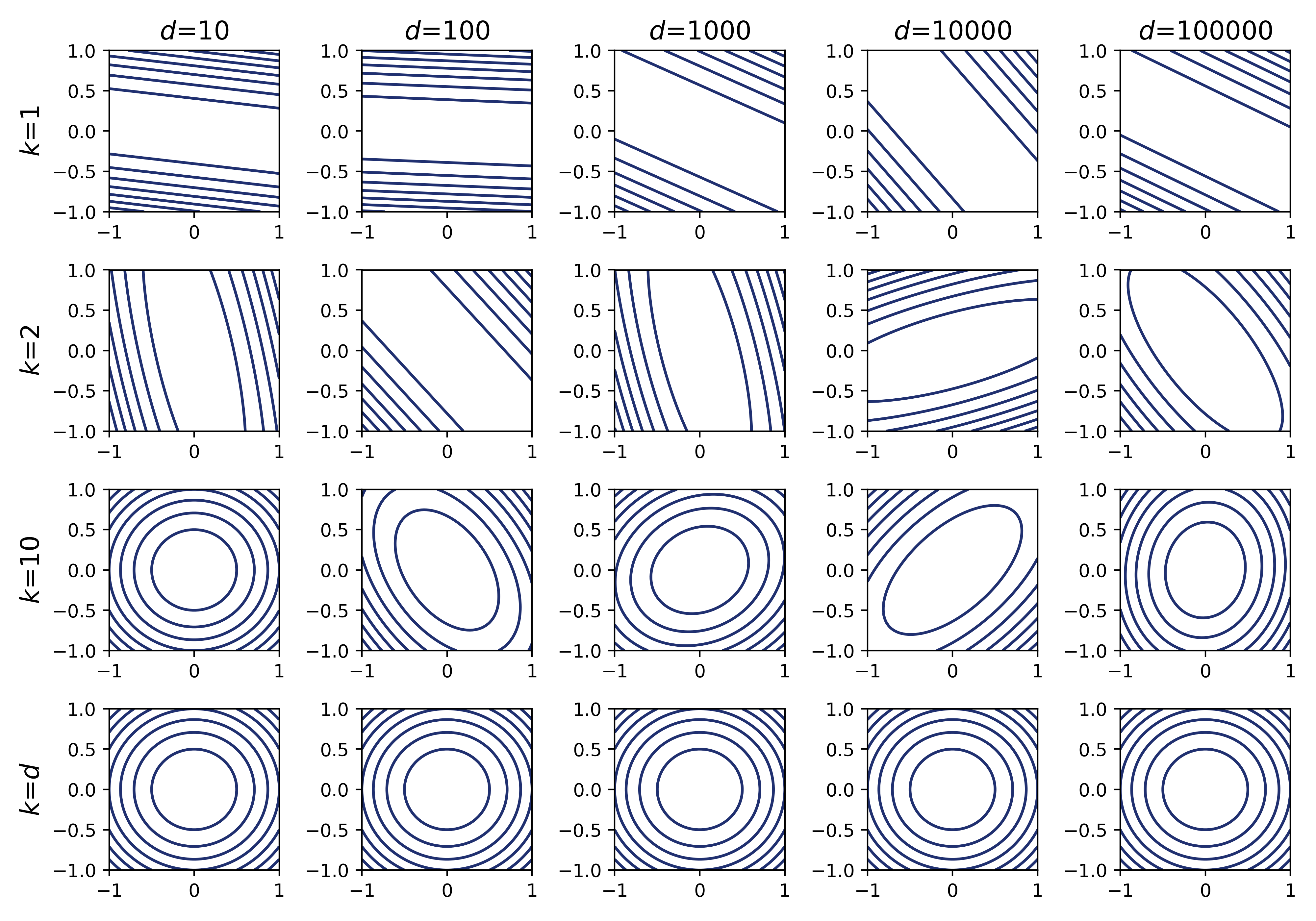}
\caption{Contour plots of random slices of $f(\mathbf{x})$ in Equation \ref{eq:kdapprox} with different $k$ and $d$. Visualized ill-conditioning is more accurate with small $k$. }\label{fig:kd}
\end{figure}

It turns out that \textit{visualization accuracy depends on the intrinsic dimensionality $k$}. To analyze this, let us consider the extreme case of $\epsilon = 0$, i.e.,  
\begin{equation}
f(\mathbf{x}) = f(\mathbf{x}_{:k}) = ||\mathbf{x}_{:k}||^2.  \label{eq:kdapprox}
\end{equation}

Fig. \ref{fig:kd} shows the contour plots of random 2D slices with different $k$ and $d$. With low $k$, the visualized ill-conditioning is more accurate, independent of full problem dimensionality $d$. Deriving a closed-form expression of $\kappa_{2D}$ as a function of $d$ and $k$ is beyond the scope of this paper. However, as shown below, $k=1$ and $k=d$ result in the correct $\kappa_{2D}=\infty$ and $\kappa_{2D}=1$, respectively. 

Let $\mathbf{u}, \mathbf{v} \in \mathbb{R}^d$ denote the slice basis vectors, with orthogonality $\mathbf{u}^T\mathbf{v}=0$. On the visualization plane, $\mathbf{x}=p_1 \mathbf{u} + p_2 \mathbf{v} = [\mathbf{u}\  \mathbf{v}]\mathbf{p}$, where $\mathbf{p}$ denotes the 2D position on the plane. Similarly, $\mathbf{x}_{:k}=[\mathbf{u}_{:k} \  \mathbf{v}_{:k}]\mathbf{p}$, and the objective can be expressed as:
\begin{eqnarray}
f(\mathbf{x}_{:k}) &=& ||[\mathbf{u}_{:k} \mathbf{v}_{:k}]\mathbf{p}||^2\\
&=& \mathbf{p}^T \begin{bmatrix}
\mathbf{u}^T_{:k}\mathbf{u}_{:k} & \mathbf{u}^T_{:k}\mathbf{v}_{:k} \\
\mathbf{v}^T_{:k}\mathbf{u}_{:k} & \mathbf{v}^T_{:k}\mathbf{v}_{:k}
\end{bmatrix} \mathbf{p} \\
&=& \mathbf{p}^T \mathbf{A} \mathbf{p}.
\end{eqnarray}

Because $\mathbf{A}$ is symmetric, the Hessian of the quadratic form w.r.t. $\mathbf{p}$ is:
\begin{eqnarray}
H(f(\mathbf{x}_{:k}))=\mathbf{A}+\mathbf{A}^T=2\mathbf{A}.
\end{eqnarray}

The condition number $\kappa_{2D}=\kappa(H(f(\mathbf{x}_{:k}))=\kappa(\mathbf{A})$, as the condition number is invariant to scaling the Hessian by a constant.  

With $k$=1, the vectors $\mathbf{u}_{:k}=[u_1], \mathbf{v}_{:k}=[v_1]$, and the determinant becomes zero:
\begin{eqnarray}
\det(\mathbf{A})&=& \mathbf{u}^T_{:k}\mathbf{u}_{:k}\mathbf{v}^T_{:k}\mathbf{v}_{:k} - \mathbf{u}^T_{:k}\mathbf{v}_{:k}\mathbf{v}^T_{:k}\mathbf{u}_{:k} \\
&=&u_1 u_1 v_1 v_1 - u_1 v_1 v_1 u_1 = 0.
\end{eqnarray}

This indicates at least one zero eigenvalue and, since $\mathbf{A}$ is not a null matrix, some eigenvalue must also be nonzero, i.e., $\kappa(\mathbf{A})=\max(eig(\mathbf{A}))/\min(eig(\mathbf{A}))=\infty$.

When $k$ grows from 1 to $d$, the vectors $\mathbf{u}_{:k}, \mathbf{v}_{:k}$ gradually become closer to $\mathbf{u}, \mathbf{v}$, i.e., unit-length and orthogonal. Thus, the off-diagonal elements become zero and $\mathbf{A}$ becomes the identity matrix, with $\kappa(\mathbf{A})=1$.

Although the quadratic $f(\mathbf{x})$ was chosen to be separable for easier mathematical analysis, the result generalizes to the arbitrarily rotated case.

\subsection{2D Visualizations Can Be Optimistic About Multimodality}
\begin{figure}[t]
\centering
\includegraphics[width=\linewidth]{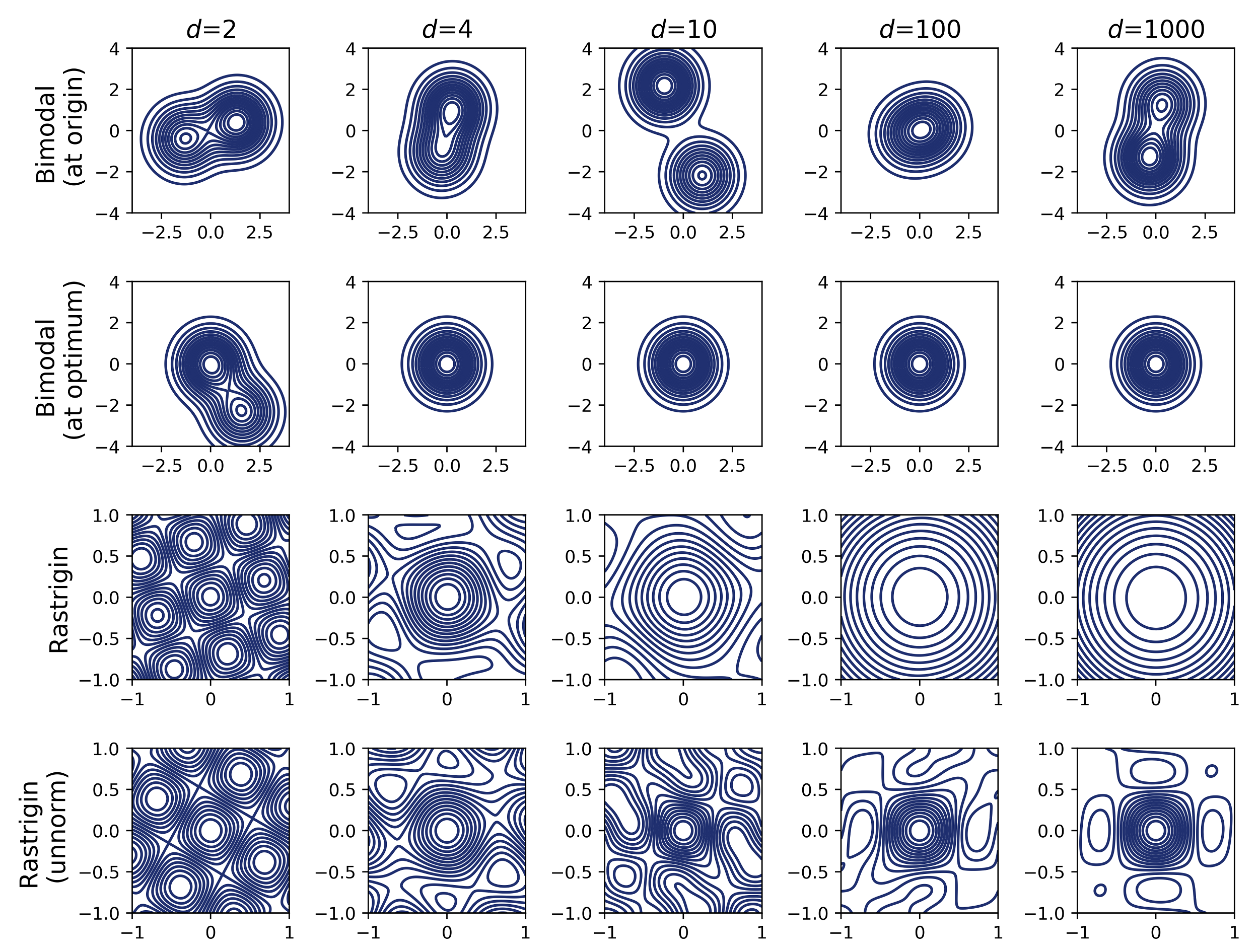}
\caption{Contour plots of random slices of multimodal test functions with different $d$. Large $d$ can make multimodality less apparent, although it can be compensated by using unnormalized visualization basis vectors (bottom row). On the first row, the visualization plane is centered at the origin, between the optima. On the second row, it is centered at the optimum.}\label{fig:nonconvex}
\end{figure}

Fig. \ref{fig:nonconvex} shows contour plots of random visualization slices with different dimensionality $d$, using two multimodal test functions. Rastrigin's function is a standard multimodal optimization test function with the global minimum at the origin and infinitely many local minima:
\begin{equation}
f_{Rastrigin}(\mathbf{x})=10d+\sum_{i=1}^d [x_i^2-10\cos(2\pi x_i)]
\end{equation}

Additionally, we use the following bimodal function:
\begin{equation}
f_{Bimodal}(\mathbf{x})=e^{-\frac{1}{2}||\mathbf{x}-\mathbf{1}||^2} + 0.8e^{-\frac{1}{2}||\mathbf{x}+\mathbf{1}||^2},
\end{equation}
where $\mathbf{1}$ denotes a vector of ones. We visualize this function both around the origin and around the dominant mode at $\mathbf{1}$. 

Fig. \ref{fig:nonconvex} reveals two key insights:
\begin{itemize}
    \item \textit{Multimodality becomes less apparent with increasing dimensionality $d$.} The exception is the first row, where visualized multimodality only depends on plane rotation independent of $d$. This is because the visualization plane intersects the origin; in this case, the optima are always equally far from the plane and thus have similar influence on the visualized $f_{Bimodal}(\mathbf{x})$. However, when the plane intersects the optimum at $\mathbf{1}$, the other optimum tends to lie increasingly far from the plane with increasing $d$, having a negligible effect on the visualization.
    \item The visualization of Rastrigin's function illustrates how \textit{landscape features may scale differently with dimensionality}. Rastrigin's central mode becomes more dominant with large $d$. At the bottom of Fig. \ref{fig:nonconvex}, we demonstrate how this can be compensated by omitting the unit-length normalization of the slice basis vectors, and instead normalizing them to the mean of their sampled lengths. Each basis vector element is sampled uniformly in the range $[-1,1]$.
\end{itemize}

\subsection{If 2D Visualizations Show Problems, There Really Are Problems}
\textbf{\textit{Visualized ill-conditioning is real}} The results above indicate that 2D visualizations have limited sensitivity as a diagnostic tool for detecting ill-conditioning and multimodality. Fortunately, $\kappa_{2D} \le \kappa$ also means that the visualizations have high specificity, i.e. if the 2D isocontours are elongated, the problem is indeed ill-conditioned. 

\textbf{\textit{Visualized non-convexity indicates real non-convexity}} For a convex unimodal objective, the 2D visualization is likewise convex and unimodal. This follows from the intersection of two convex sets being convex. Each 2D isocontour encloses a set that is the intersection of two convex subsets of $\mathbb{R}^d$, i.e., the visualization plane and the volume enclosed by the corresponding isosurface. However, other non-convexity can be confused with multimodality. Consider a curved, banana-shaped 3D isosurface. It is possible to intersect this with a plane such that the resulting isocontours comprise two ellipses.

\subsection{Summary of Limitations}
In summary, random 2D visualization slices of high-dimensional objectives tend to be optimistic about both ill-conditioning and multimodality. However, this limitation is mitigated by the visualizations not showing illusory non-convexity or ill-conditioning. In other words, \textit{as a diagnostic tool for detecting problems, random 2D visualizations have low sensitivity compensated by high specificity}. 

Mitigating the low sensitivity is a potential topic for future work. For instance, if computing eigenvectors and eigenvalues of the Hessian or its low-rank approximation (e.g., \cite{li1992principal}) is not too expensive, one could visualize using a 2D basis formed by the eigenvectors with lowest and highest eigenvalues. This way, the visualization would be in line with the condition number and show ill-conditioning with higher sensitivity. In this paper, however, we have focused on random visualization slices due to their simplicity and prior success in visualizing neural network loss landscapes \cite{li2018visualizing}. Furthermore, at least for large policy networks with millions of parameters, computing eigenvector approximations is quite expensive. %It could also be possible to use numerical optimization to find directions of (approximately) highest and lowest landscape curvature, i.e., approximations of only the first and last eigenvectors.